\documentclass{article}

\usepackage[numbers]{natbib}
\usepackage[preprint]{neurips_2026}

\usepackage[utf8]{inputenc} %
\usepackage[T1]{fontenc}    %
\usepackage{hyperref}       %
\usepackage{url}            %
\usepackage{booktabs}       %
\usepackage{amsfonts}       %
\usepackage{nicefrac}       %
\usepackage{microtype}      %
\usepackage{xcolor}         %
\usepackage{amsmath}
\usepackage{amssymb}
\usepackage[dvipsnames]{xcolor}
\usepackage{graphicx} %
\usepackage{caption}
\usepackage{subcaption}
\usepackage{hyperref}
\usepackage{booktabs}
\usepackage{array}
\usepackage{color,soul}
\usepackage{ulem} %
\usepackage{subcaption}
\usepackage{wrapfig}
\usepackage[most]{tcolorbox}
\usepackage{enumitem}

\definecolor{accentcolor}{RGB}{40, 70, 120}

\newcommand{\akarsh}[1]{{\color{red}AK: #1}}
\newcommand{\akarshdel}[1]{{}}

\title{Pretraining Recurrent Networks without Recurrence}

\author{%
Akarsh Kumar
\qquad \qquad
Phillip Isola \\\\
\quad \quad MIT
}

\begin{document}

\maketitle

\begin{figure}[h!]
    \centering
    \vspace{-25pt}
    \includegraphics[width=1.0\linewidth]{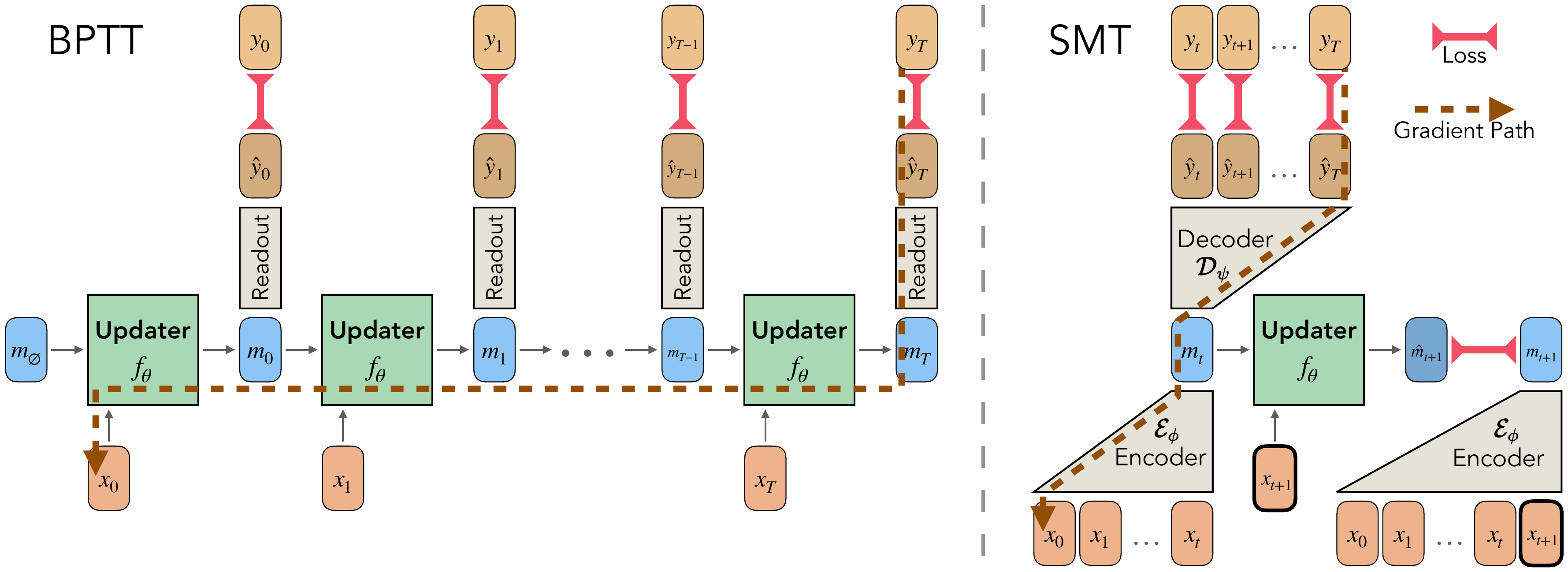}
    \caption{
    \textbf{BPTT vs SMT.}
    \textbf{Left:}
    BPTT trains an RNN by recurrently unrolling the ``updater'' network in time, and backpropagating gradients through the entire graph.
    \textbf{Right:}
    Supervised Memory Training (SMT) trains an RNN with supervised learning on one-step memory transition labels, which are generated by a Transformer encoder-decoder model pair trained to produce \textit{predictive states}.
    SMT is fully time-parallel.
    In SMT, the longest gradient path between tokens is $\mathcal O(1)$ (compared to $\mathcal O(T)$ in BPTT), which stabilizes gradients, making learning long-range dependencies qualitatively easier. 
    }
    \label{fig:teaser}
\end{figure}

\begin{abstract}
Training recurrent neural networks (RNNs) requires assigning credit across long sequences of computations.
Standard backpropagation through time (BPTT) addresses this problem poorly:
it is sequential in time, limiting parallelism, and suffers from vanishing or exploding gradients, making long-range associations difficult to learn.
We propose \textit{Supervised Memory Training} (SMT), a method for training nonlinear RNNs that sidesteps recurrent credit propagation entirely by reducing RNN training to supervised learning on one-step memory transition labels $(m_t, x_{t+1}) \rightarrow m_{t+1}$.
SMT acquires these memory labels by training a Transformer-based encoder on a predictive state objective---retaining only information from the past necessary to predict the future.
By decoupling \textit{what to remember} from \textit{how to update memory}, SMT enables time-parallel RNN training with a stable $\mathcal{O}(1)$ length gradient path between any two tokens---without ever unrolling the RNN.
We find that SMT outperforms BPTT when pretraining various RNN architectures on tasks like language modeling and pixel sequence modeling.
SMT enables nonlinear RNNs to better capture long-range dependencies and train in parallel, potentially unlocking the scaling of models that build temporal abstractions of past experience.

\begin{tcolorbox}[
    enhanced,
    boxrule=0pt,
    frame hidden,
    left=5pt,
    right=5pt,
    top=4pt,
    bottom=4pt
]
\textsf{\textbf{Project Page:}} \href{https://akarshkumar.com/smt}{\texttt{akarshkumar.com/smt}} \\
\textsf{\textbf{Source Code:}} \href{https://github.com/akarshkumar0101/smt}{\texttt{github.com/akarshkumar0101/smt}}
\end{tcolorbox}

\end{abstract}

\newpage

\section{Introduction}
\label{sec:intro}

\vspace{-8pt}
Recurrent neural networks (RNNs) store information about the past that will only become useful in the future.
The core training challenge is that the utility of a memory may be delayed:
many intermediate computations intervene between writing information and eventually using it.
These intervening steps confound learning the correct associations, a problem known as \textit{credit assignment}~\citep{minsky1961steps}.

The standard approach, backpropagation through time (BPTT), assigns credit across a sequence by unrolling the RNN in time and propagating gradients backward through the resulting computation graph~\citep{rumelhart1986learning,werbos1990backpropagation}.
Although conceptually well-motivated, BPTT is sequential in time and suffers from unstable high variance gradients that may vanish or explode~\citep{pascanu2013difficulty}.
The lack of time-parallelism makes BPTT scale poorly, while its gradient instability makes learning long-range associations difficult, as credit must propagate across up to $\mathcal{O}(T)$ steps~\citep{bengio1994learning}.
Is recurrent credit propagation unavoidable?

In this paper, we propose \textit{Supervised Memory Training} (SMT), a method to train nonlinear RNNs that sidesteps recurrent credit propagation by reducing the problem to supervised learning.
Suppose we had access to the optimal memory state at each timestep, $m^*_t$.
Then, RNN training reduces to learning the one-step update $(m^*_t, x_{t+1})\rightarrow m^*_{t+1}$ using standard supervised objectives.

The challenge, of course, is how to actually obtain such memory labels.
In this paper, we assert that an effective memory is a sufficient statistic of the past for predicting the future, i.e., a predictive state~\citep{littman2001predictive}.
The past is typically viewed as a sequence, suggesting that memory must be computed sequentially over time.
Our key insight is that, by augmenting each observation with its timestamp, the past can instead be losslessly represented as a \textit{set} of timestamped events, rather than a sequence.
Under this reparameterization, the optimal memory becomes a permutation-invariant function of this set, and can therefore be estimated using models that operate in parallel over time. %
This reframing allows us to train memory representations \textit{without recurrently propagating credit through time}.

In practice, we train a Transformer encoder model to embed the past context into a memory that a separate decoder can use to predict the future.
This objective operationalizes the notion of a predictive state: a representation of the past that retains only the information needed to predict the future and nothing more.
Once this teacher encoder has learned to construct such memory representations, the RNN can then focus on learning the now much simpler task of updating that memory over time.

In essence, SMT decouples learning \textit{what to remember} (memory representation), which is a non-sequential problem, from learning \textit{how to update memory} (memory dynamics), which is a sequential process but can be supervised one-step at a time.
This decoupling enables time-parallel training of nonlinear RNNs \textit{without unrolling}, and creates a stable $\mathcal{O}(1)$ gradient path for long-range associations.

Indeed, Transformers solved time-parallelism and credit assignment in the same way~\citep{vaswani2017attention}, and have since revolutionized sequence modeling~\citep{brown2020language}.
However, Transformers do not possess a compressed memory of the past in the way RNNs or human brains do~\citep{gu2025tradeoffs,kandel2000principles}.
Instead, Transformers store the \textit{entire history} of past token representations and attend to \textit{all of them} when processing each new token.
As a result, their memory size grows with sequence length, leading to prohibitive computational costs for unbounded sequences, such as a human lifetime of experience~\citep{tay2022efficient}.
Sliding-window transformers mitigate this issue by storing only the most recent tokens, but have the severe drawback that they lose access to information before the context window~\citep{dai2019transformer}.
In contrast, no known biological intelligence operates in this manner---accessing its entire experiential history for every new decision---but instead constructs a temporally compressed abstraction of past experience, like an RNN~\citep{bennett2023brief}.

Linear attention RNN models also exhibit time-parallel training and relatively stable credit assignment, while maintaining a fixed memory size~\citep{katharopoulos2020transformers,gu2021efficiently,gu2023mamba,dao2024transformers}.
But, because their transition function is linear, the class of functions they can represent is fundamentally constrained~\citep{merrill2026linear}, which can lead to failure on important sequential tasks such as state tracking~\citep{merrill2024illusion,liu2025serial}.

SMT aims to combine the best of all worlds:
time-parallel training, stable $\mathcal O(1)$ long-range credit assignment, fixed-memory inference,
and maximal expressivity via nonlinear dynamics.
Our results confirm that, on language modeling and pixel sequence modeling tasks, SMT outperforms BPTT in learning long-range dependencies while requiring less sequential computation.
SMT should primarily be used for pretraining RNNs, followed by some lightweight post-training to mitigate drift from the teacher memory trajectories and adapt to specific downstream tasks.
In fact, post-training is necessary to go beyond the limitations of the teacher encoder~\citep{merrill2023parallelism}.
Beyond its role as a training approach for RNNs, SMT can also be seen as a new method for learning representations (mappings from data to latent variables) and for learning world models (transitions from state at time $t$ to state at time $t+1$).

\section{Methods}
\label{sec:methods}

\vspace{-5pt}
\subsection{Background}
\label{sec:methods_background}
\vspace{-5pt}

\paragraph{Causal Conditional Sequence Modeling}
Let $\mathbf{x} = [x_0, \dots, x_T]$ and $\mathbf{y} = [y_0, \dots, y_T]$ denote input and output sequences.
The objective is to learn a model of the conditional distribution $p(\mathbf{y} \mid \mathbf{x})$.
We assume each output $y_t$ depends only on $x_0,\dots,x_t$.
Formally we model this distribution with $\prod_{t=0}^{T} p_\theta(y_t \mid \mathbf{x}_{\leq t})$.
Autoregressive sequence modeling is a special case when $x_t = y_{t-1}$.

\vspace{-5pt}
\paragraph{Recurrent Neural Networks (RNNs)}
An RNN models this problem using a fixed-size latent state, $m_t$, that summarizes past inputs.
At each timestep, this state is updated according to:
\begin{equation}
\label{eq:rnn_update}
m_{t+1} = f_\theta(m_t, x_{t+1})
\end{equation}
where $f_\theta$ is the transition function.
The predicted output token distribution is then
$ p_\theta(y_{t} \mid \mathbf{x}_{\leq t}) = \text{softmax}(g_\theta(m_t)) $,
where $g_\theta$ is the readout function.
Ideally, $m_t$ ``remembers'' important information from the past and intentionally ``forgets'' unimportant information, i.e., $m_t$ is a \textit{memory}.

\vspace{-5pt}
\paragraph{Backpropagation Through Time (BPTT)}
Traditionally, RNNs are trained with BPTT~\citep{rumelhart1986learning,werbos1990backpropagation}. 
In the forward pass, $f_\theta$ is recurrently unrolled over the sequence.
The input sequence $x_t$ is provided via teacher forcing, while the memory sequence $m_t$ is generated by the RNN’s transition and is used to compute the output predictions.
Conceptually, the computation graph takes the form:
\[
m_t = f_\theta(\ldots f_\theta(f_\theta(m_{\emptyset}, x_0), x_1), \ldots , x_t) \quad\text{with}\quad m_{\emptyset}=\mathbf{0}
\]
Gradients are then computed end-to-end on this unrolled computation graph, propagating from the output prediction losses backward through the trajectory of the nonlinear dynamical system.
Thus, this gradient credit assignment signal may have to travel for a path length of up to $\mathcal{O}(T)$ steps.
Depending on the singular values of the Jacobian of $f_\theta$, gradients may vanish or explode in time.

BPTT has two well-known limitations:
\begin{enumerate}
    \vspace{-5pt}
    \item Equation~\ref{eq:rnn_update} is usually implemented with a recurrent for-loop, preventing parallelization~\citep{pascanu2013construct}.
    \item BPTT often produces unstable high variance gradients~\citep{bengio1994learning}.
    When gradients vanish, the RNN experiences a recency bias, hindering the learning of long-range associations~\citep{ravfogel2019studying}.
    When gradients explode, the induced dynamical system is chaotic, causing training instability~\citep{pascanu2013difficulty}.
\end{enumerate}

\vspace{-10pt}
\subsection{Supervised Memory Training (SMT)}
\label{sec:smt}
\vspace{-5pt}
We propose \textit{Supervised Memory Training} (SMT) for pretraining nonlinear RNNs without BPTT.
The core idea is to decouple the learning of memory representation from memory dynamics.

\vspace{-5pt}
\paragraph{Motivation}

Consider a hypothetical oracle memory-encoding model $\mathcal{Q}$, that takes as input the sequence of tokens up to timestep $t$, $\mathbf{x}_t^\text{ctx}=[x_0,x_1,\dots,x_t]$, and outputs an effective compressed memory for that timestep $m_t^*=\mathcal{Q}(\mathbf{x}_t^\text{ctx})$.
This memory retains all information from the past input that is relevant for predicting the future output, $\mathbf{y}_t^\text{fut}=[y_{t},\dots,y_T]$, while deliberately discarding unimportant details.
For example, the oracle would remember the personalities of characters in a story, but discard details of what they were wearing on a specific day, just as humans do.
Running $\mathcal Q$ at different points along the sequence produces a corresponding sequence of memory labels $[m_0^*, m_1^*, \dots, m_T^*]$.
With $\mathcal Q$, the RNN's problem of learning a temporal update collapses to standard supervised learning on oracle memory transitions labels $(m_t^*, x_{t+1})\rightarrow m_{t+1}^*$.
Our key insight is that $\mathcal{Q}$ does \textit{not} have to be a recurrent function over $[x_0, x_1, \dots, x_t]$, but can instead be represented as a permutation-invariant function over the \textit{set} $\{(x_0, 0), (x_1, 1), \dots, (x_t, t)\}$ (details in Appendix~\ref{sec:set_reframing}).

In practice, SMT approximates $\mathcal Q$ by training a time-parallel model (e.g. a Transformer) to compress the past input into a memory representation that a separate decoder model can use to predict the future output.
This future-predicting objective operationalizes the notion of a predictive state~\citep{littman2001predictive}.

\vspace{-5pt}
\paragraph{Formulation}

Formally, we have the RNN $f_\theta$, bidirectional encoder $\mathcal{E}_\phi$, and causal decoder $\mathcal{D}_\psi$.
$\mathcal E_\phi$ and $\mathcal D_\psi$ are time-parallel Transformer architectures.
Given $\mathbf{x} = [x_0, x_1, \dots, x_T]$ and $\mathbf{y} = [y_0, y_1, \dots, y_T]$, we consider, for each timestep $t$, a decomposition into the past and future:
\[
\mathbf{x}_t^{\text{ctx}} = [x_0, \dots, x_t] \qquad
\mathbf{x}_t^{\text{fut}} = [x_{t+1}, \dots, x_T] \qquad
\mathbf{y}_t^{\text{fut}} = [y_{t}, \dots, y_T]
\]
The encoder maps each context to a memory state with $m_t = \mathcal E_\phi(\mathbf{x}_t^{\text{ctx}})$.
Then, the decoder predicts the future output distribution using the memory of the past and teacher forced \textit{future inputs}:
\[
p_{\phi,\psi}(\mathbf{y}_t^\text{fut} \mid \mathbf{x}_t^\text{ctx}, \mathbf{x}_t^\text{fut})
= \prod_{\tau=t}^{T} p_\psi(y_{\tau} \mid m_t, \mathbf{x}_{t+1:\tau})
= \mathcal D_\psi(m_t, \mathbf{x}_t^\text{fut})
\]
The future decoding loss for timestep $t$ is ($\mathrm{CE}$ denotes the sequence level cross-entropy loss):
\begin{equation}
\label{eq:loss_dec}
\mathcal{L}_t^\text{dec} =
\mathrm{CE}\left(
\mathbf{y}_t^\text{fut},
p_{\phi,\psi}(\mathbf{y}_t^\text{fut} \mid \mathbf{x}_t^\text{ctx}, \mathbf{x}_t^\text{fut})
\right)
\end{equation}

We have the RNN predict the next memory given the current memory and the next input with
$\hat{m}_{t+1} = f_\theta(m_t, x_{t+1})$.
This prediction is supervised with the next timestep's memory:
\begin{equation}
\label{eq:loss_dyn}
\mathcal{L}^\text{dyn}_t = \mathrm{MSE} (\hat{m}_{t+1}, m_{t+1})
\end{equation}
This dynamics loss has two distinct purposes: 1) to train the RNN and 2) to explicitly shape the encoder memory representations to be \textit{Markovian}  (i.e. $m_{t+1}$ is predictable solely from $(m_t, x_{t+1})$).

We add a uniformity loss~\citep{wang2020understanding} to prevent the memory space from collapsing:
\begin{equation}
\mathcal{L}^\text{unif} = \log \mathbb{E}_{t_a, t_b \sim [0,\dots,T]} \exp(-2\| m_{t_a} - m_{t_b}\|_2^2)
\end{equation}

The full objective is a weighted sum of all three losses:
\begin{equation}
\label{eq:smt}
\mathcal{L}^\text{smt} =
\lambda_\text{dec} \mathbb{E}_t\left[ \mathcal{L}^{\text{dec}}_t \right]
+ \lambda_\text{dyn} \mathbb{E}_t\left[ \mathcal{L}^{\text{dyn}}_t \right]
+ \lambda_\text{unif} \mathcal{L}^\text{unif}
\end{equation}
where the $\lambda$ terms control the trade-off between memory representation, dynamics, and collapse.

\vspace{-5pt}
\paragraph{Practice}
Theoretically, it should be enough to train $\mathcal{E}_\phi$ and $\mathcal D_\psi$ with only $\mathcal{L}^{\text{dec}}$, and separately train $f_\theta$ with only $\mathcal{L}^{\text{dyn}}$ (proof in Appendix~\ref{sec:markovian_encoder_proof}).
However, in practice we find it beneficial to jointly train all models in one stage with $\mathcal{L}^{\text{smt}}$, since that explicitly optimizes $m_t$ to be Markovian, and provides additional temporal credit propagation benefits described in Section~\ref{sec:detach_rnn_experiments}.

For experiments, we truncate $\mathbf{x}_t^\text{ctx}$ to a context length $T_c$ and $\mathbf{y}_t^\text{fut}$ to a future length $T_f$.
For computational efficiency, we estimate the expectation in $\mathcal{L}^{\text{smt}}$ by randomly sampling a single timestep $t$, rather than computing all timesteps in the sequence.
This yields SMT a smaller training memory footprint than BPTT: $\mathcal{O}(M+T)$ instead of $\mathcal{O}(MT)$, where $M$ is the memory size.

\vspace{-5pt}
\paragraph{Properties of SMT}
In SMT, the encoder model constructs appropriate memory representations of the past, while the RNN is responsible for learning the now much simpler task of updating that memory in one-step, thereby decoupling memory representation from memory dynamics.
In contrast, under BPTT training, the RNN must learn both tasks simultaneously.
Since the memory labels are acquired with a ``teacher'' encoder-decoder pair, SMT inherits all of its properties, such as time-parallelism, $\mathcal{O}(1)$ credit path for long-range associations, and gradient stability.

\begin{wrapfigure}{r}{0.45\textwidth}
    \centering
    \vspace{-15pt}
    \includegraphics[width=0.44\textwidth]{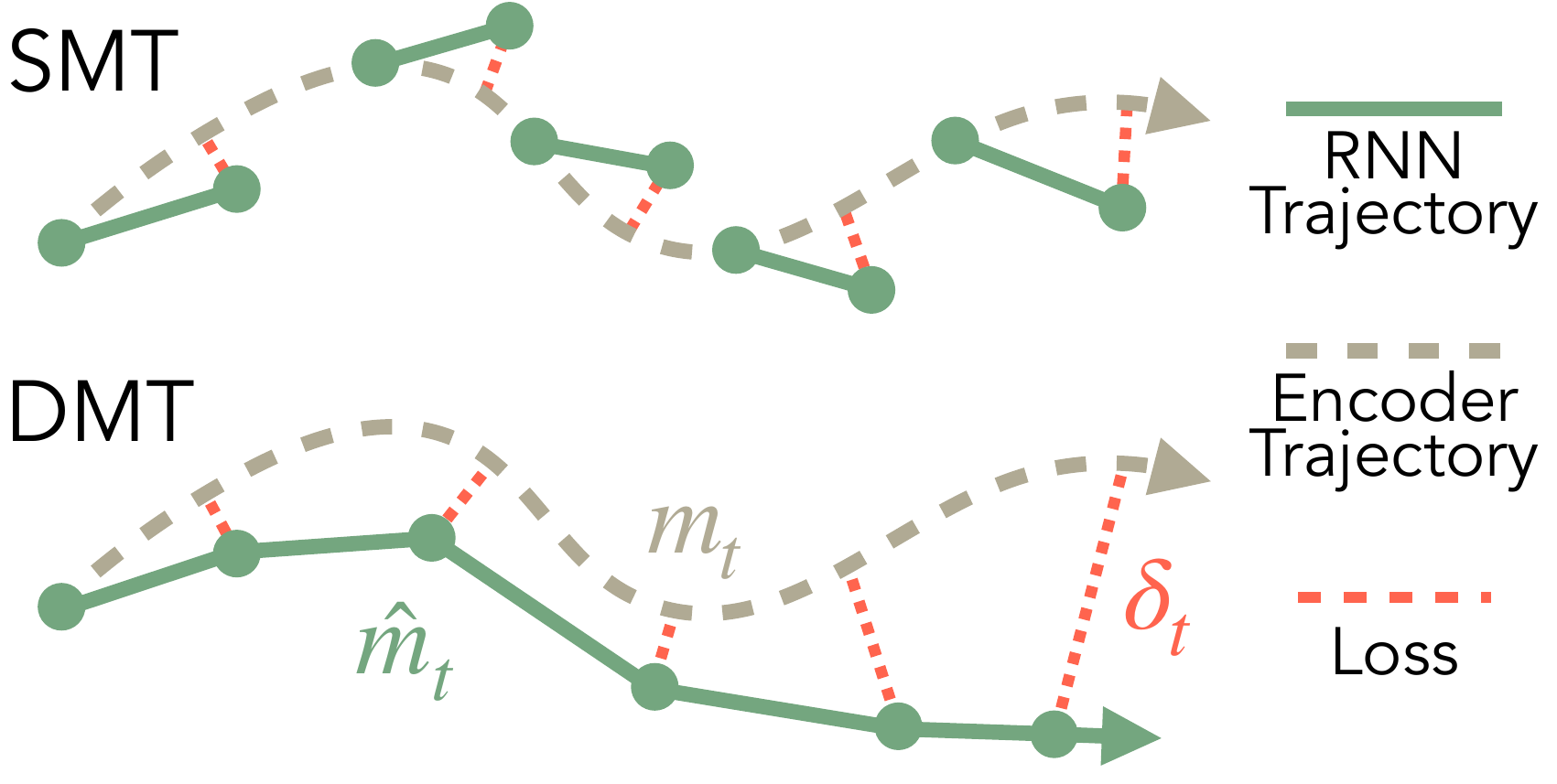}
    \caption{
    \label{fig:dmt}
    \textbf{SMT vs DMT.}
    SMT trains the RNN with behavior cloning on the encoder-generated memory states (off-policy imitation learning).
    DMT unrolls the RNN with its own memory states and then imitates the encoder trajectory (on-policy imitation learning).
    Figure design inspired by~\citet{jacobs2025block}.
    }
\end{wrapfigure}

\vspace{-5pt}
\subsection{DAgger Memory Training (DMT)}
\label{sec:dmt}
\vspace{-5pt}

After SMT, the RNN achieves low one-step error in predicting $(m_t, x_{t+1})\rightarrow m_{t+1}$ when $m_t$ comes from the encoder.
However, at evaluation time, the model is unrolled autoregressively, using its own predicted memories rather than the encoder memories as input.
This train–test mismatch causes small prediction errors to accumulate over time, leading to a growing drift between the RNN-generated memory trajectory $[\hat{m}_0, \dots, \hat{m}_T]$ and the encoder trajectory $[m_0, \dots, m_T]$, even with teacher forced input tokens.
This drift is quantified as $\delta_t = \text{MSE}(\hat m_t, m_t)$.

We introduce \textit{DAgger Memory Training} (DMT), a finetuning phase that corrects this drift via \textit{on-policy imitation learning}~\citep{ross2011reductionimitationlearningstructured}.
By exposing the RNN to its own induced memory state distribution, DMT trains the RNN to autocorrect its errors to stay aligned with the encoder trajectory (Figure~\ref{fig:dmt}).

Concretely, given $\mathbf{x}$, we first compute the encoder trajectory $[m_0, \dots, m_T]$ using only $\mathcal E_\phi$ and then the RNN trajectory $[\hat m_0, \dots, \hat m_T]$ using $f_\theta$.
Instead of training on SMT labels $(m_t, x_{t+1}) \rightarrow m_{t+1}$, we train on DMT labels $(\hat m_t, x_{t+1}) \rightarrow m_{t+1}$.
Equivalently, the training loss is:
\begin{equation}
\mathcal{L}^\text{dmt}=\mathbb{E}_t \left[ \mathrm{MSE}(\hat{m}_t, m_t) \right]
\end{equation}

During DMT, we freeze the encoder and decoder and only train the RNN with a small learning rate.
Note that DMT unrolls the RNN memories, but still uses teacher forced $x_t$ inputs.
Although DMT unrolls the RNN and gradients may optionally propagate through time, its objective is fundamentally different than standard BPTT, since long-range credit is already assigned in the encoder memory labels, $m_t$.
DMT is not time-parallel.
That said, DMT should primarily be viewed as a lightweight fine-tuning phase following SMT.
Table~\ref{tab:resources_bigo} shows the resource requirements for the different methods.

\vspace{-12pt}
\begin{table}[h!]
\centering
\caption{
\textbf{Resource requirements.}
$T$ is token sequence length.
$T_c$ is SMT encoder context length.
For RNNs, $M$ is the memory state size.
We ignore $\log$ terms for simplicity.
LA denotes linear attention (in its parallel and recurrent form).
Complexity classes are from~\citet{merrill2026linear}.
}
\label{tab:resources_bigo}
\resizebox{\textwidth}{!}{%
\begin{tabular}{lccccc|cc|c}
\toprule
& \multicolumn{4}{c}{\shortstack{Training ($T$-Length Sequence)}} & & \multicolumn{2}{c}{\shortstack{Inference (One-Step)}} & \shortstack{Complexity Class} \\
\cmidrule(lr){2-5} \cmidrule(lr){7-8} \cmidrule(lr){9-9}
Method & Memory & Compute & \shortstack{\textbf{Sequential} \\ \textbf{Operations}} & \shortstack{\textbf{Credit Path} \\ \textbf{Length}} & & Memory & Compute & \\
\midrule
Transformer      & $\mathcal{O}(T)$      & $\mathcal{O}(T^2)$    & $\mathcal{O}(1)$ & $\mathcal{O}(1)$ & & $\mathcal{O}(T)$ & $\mathcal{O}(T)$ & $\mathrm{TC}^0$ \\
LA (parallel) & $\mathcal{O}(T)$       & $\mathcal{O}(T^2)$       & $\mathcal{O}(1)$ & $\mathcal{O}(1)$ & & $\mathcal{O}(M)$ & $\mathcal{O}(1)$ & $\mathrm{PNC}^1$ \\
LA (recurrent) & $\mathcal{O}(M)$       & $\mathcal{O}(T)$       & $\mathcal{O}(1)$ & $\mathcal{O}(1)$ & & $\mathcal{O}(M)$ & $\mathcal{O}(1)$ & $\mathrm{PNC}^1$ \\
BPTT (RNN)       & $\mathcal{O}(MT)$     & $\mathcal{O}(T)$      & $\mathcal{O}(T)$ & $\mathcal{O}(T)$ & & $\mathcal{O}(M)$ & $\mathcal{O}(1)$ & $\mathrm{L/P}$ \\
\textbf{SMT} (RNN)        & $\mathcal{O}(M+T_c)$ & $\mathcal{O}(T_c^2)$ & $\mathcal{O}(1)$ & $\mathcal{O}(1)$ & & $\mathcal{O}(M)$ & $\mathcal{O}(1)$ & $\mathrm{L/P}$ \\
\textbf{DMT} (RNN)        & $\mathcal{O}(MT)$  & $\mathcal{O}(T_c^2T)$  & $\mathcal{O}(T)$  & $\mathcal{O}(1)$ & & $\mathcal{O}(M)$ & $\mathcal{O}(1)$ & $\mathrm{L/P}$ \\
\bottomrule
\end{tabular}%
}
\end{table}

\vspace{-12pt}
\section{Experiments}
\label{sec:experiments}
\vspace{-5pt}

We study the properties of SMT and compare against BPTT, the standard RNN training algorithm.
We restrict our analysis to nonlinear RNNs, the primary setting BPTT is applied to.
Transformers and linear RNNs are excluded as they are qualitatively distinct model classes~\citep{merrill2026linear,gu2025tradeoffs}.
\vspace{-2pt}

``BPTT RNN'' denotes the BPTT baseline.
``SMT Encoder$^*$'' generates memories $m_t$ with the SMT-trained encoder and predicts next tokens using the decoder.
This method is essentially a Transformer baseline with the same memory bottleneck as our RNNs.
Since it serves as the teacher during SMT and DMT, it provides a reference upper bound on RNN performance.
``SMT$\rightarrow$DMT RNN'' denotes the RNN pretrained with SMT and finetuned with DMT, which constitutes our full method.

\vspace{-8pt}
\paragraph{Architectures}
We test RNNs with a Transformer~\citep{vaswani2017attention}, MLP, GRU-gated~\citep{cho2014properties}, and LSTM-gated~\citep{hochreiter1997long} backbone.
In Appendix~\ref{sec:additional_experiments}, we also evaluate vanilla stacked GRUs and LSTMs.
\vspace{-8pt}
\paragraph{Datasets}
We consider character-level language modeling on TinyStories~\citep{eldan2023tinystories} as a naturalistic task requiring long-range memory~\citep{sutskever2011generating}.
As a more challenging problem, we test our method on raster-scan order pixel sequence modeling of sparse images from MNIST~\citep{lecun1998mnist} and Sketchy~\citep{sangkloy2016sketchy}.
This is a hard problem for RNNs~\citep{lamb2016professor, van2016pixel}.
Imagine you are an ant traversing an image pixel by pixel, row by row.
When you see a new white pixel, in order to recognize the shape and slope of the stroke it belongs to, you must remember the white pixels you saw in the previous rows, which may be hundreds of timesteps ago, buried among black pixels.
RNNs must achieve this with finite memory, meaning no direct attention to earlier pixels, and thus forcing long-range memory to emerge.
We term this ``Attneave's task'', based on classic work from perceptual psychology~\citep{attneave1954some}.
\vspace{-4pt}

More details on architectures, datasets, and experiments are in Appendix~\ref{sec:appendix_experiment_details}.

\vspace{-10pt}
\subsection{Synthetic Task Experiments}
\label{sec:synthetic_experiments}
\vspace{-8pt}

\begin{figure}[t]
    \centering
    \includegraphics[width=1.0\linewidth]{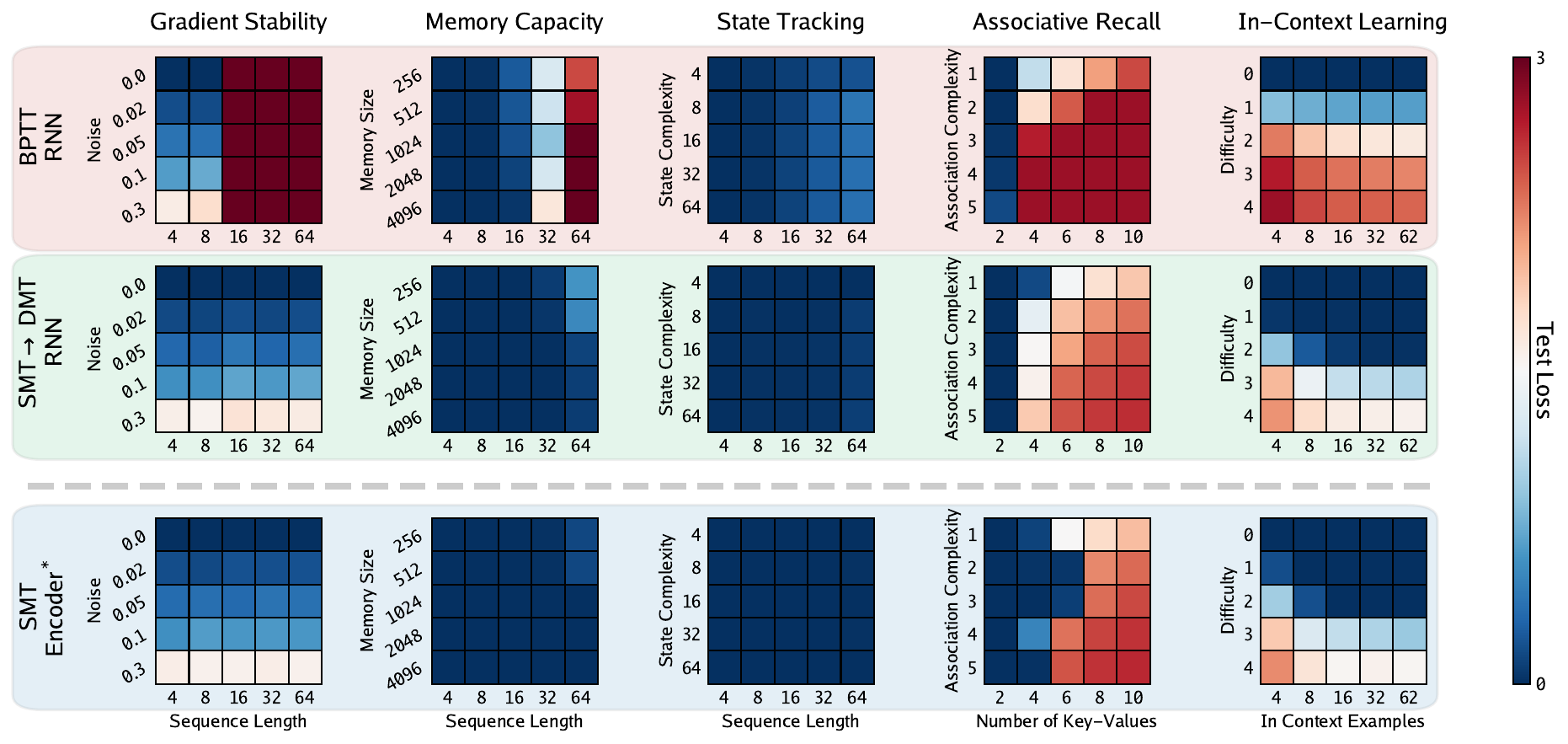}
    \caption{
    \textbf{Synthetic Task Experiments.}
    We evaluate BPTT, SMT, and SMT$\rightarrow$DMT using five synthetic tasks with various settings to probe different properties of the algorithms.
    $^*$ signifies that the SMT Encoder is the teacher Transformer (\textit{not} an RNN) and is used only as a reference.
    Across all tasks and task settings, SMT$\rightarrow$DMT outperforms BPTT, signaling that SMT has better gradient properties, memory utilization, state tracking, associative recall, and in-context learning than BPTT.
    }
    \label{fig:synthetic_results}
    \vspace{-18pt}
\end{figure}

We first evaluate BPTT and SMT on synthetic tasks designed to isolate and probe specific properties of the training algorithm.
The RNN architecture with the Transformer backbone is used for these experiments.
For these synthetic experiments, we set $T_c=T_f=T$ and train all timesteps in the $\mathcal{L}^\text{smt}$ expected value.
Our tasks include the following
(details of tasks are in Appendix~\ref{sec:synth_task_desc}):
\begin{enumerate}
    \vspace{-8pt}
    \item Retrieval to test \textbf{Gradient Stability} (sweep sequence length and noise level).
    \vspace{-4pt}
    \item String Copy to test \textbf{Memory Capacity} (sweep sequence length and memory state size).
    \vspace{-4pt}
    \item  Stack Operations to test \textbf{State Tracking} (sweep sequence length and state complexity).
    \vspace{-4pt}
    \item Keys-Values to test \textbf{Associative Recall} (sweep number of and complexity of associations).
    \vspace{-4pt}
    \item Modular Arithmetic to test \textbf{In-Context Learning} (sweep difficulty and number of examples).
\end{enumerate}
\vspace{-4 pt}

Figure~\ref{fig:synthetic_results} shows that SMT$\rightarrow$DMT outperforms BPTT in all settings of all tasks.
BPTT struggles to learn as sequences get longer, even when the task is simple, e.g. retrieval.
It also struggles to utilize memory capacity fully, do associative recall, and perform in-context learning, all of which require solving long-range credit assignment.
In contrast, SMT seems agnostic to the sequence length, and is able to solve all of the harder credit assignment problems except associative recall.
We attribute these differences to BPTT's $\mathcal{O}(T)$ credit path length, compared to SMT's $\mathcal{O}(1)$.
Further analysis in Section~\ref{sec:vanishing_grads} confirms the difference in gradient stability in both methods.

\newpage

\vspace{-5pt}
\subsection{Attneave's Pixel Sequence Modeling}
\label{sec:attneaves_experiments}
\vspace{-10pt}

\begin{wrapfigure}{r}{0.5\textwidth}
  \centering
  \vspace{-25pt}
  \begin{minipage}{\linewidth}
    \centering
    \includegraphics[width=\linewidth]{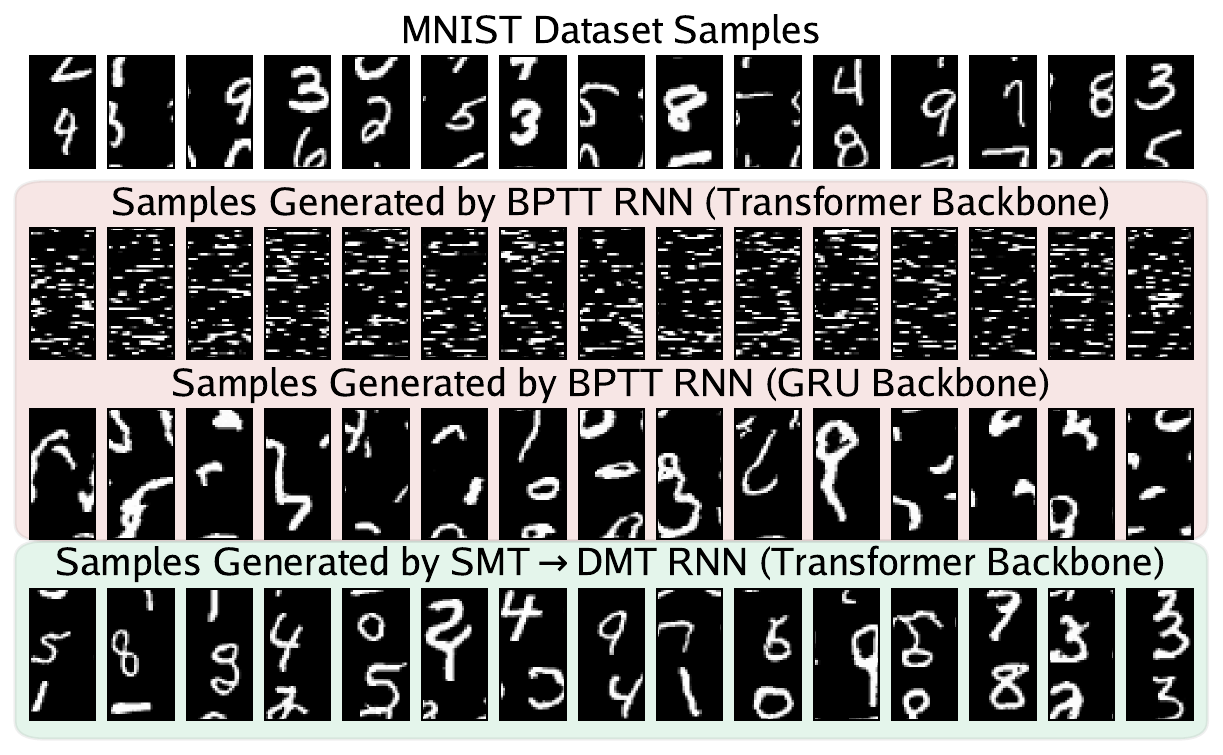}
    \captionof{figure}{%
      \textbf{Attneave's MNIST Generation.}
      BPTT fails to effectively capture the long-range dependencies required for pixel sequence modeling, even with a GRU.
      SMT$\rightarrow$DMT captures these dependencies with a non-gated RNN architecture.
      More samples are in Appendix Figure~\ref{fig:mnist_generations_extra}.
    }
    \label{fig:mnist_generations}
  \end{minipage}

  \vspace{3pt}

  \begin{minipage}{\linewidth}
    \centering
    \includegraphics[width=\linewidth]{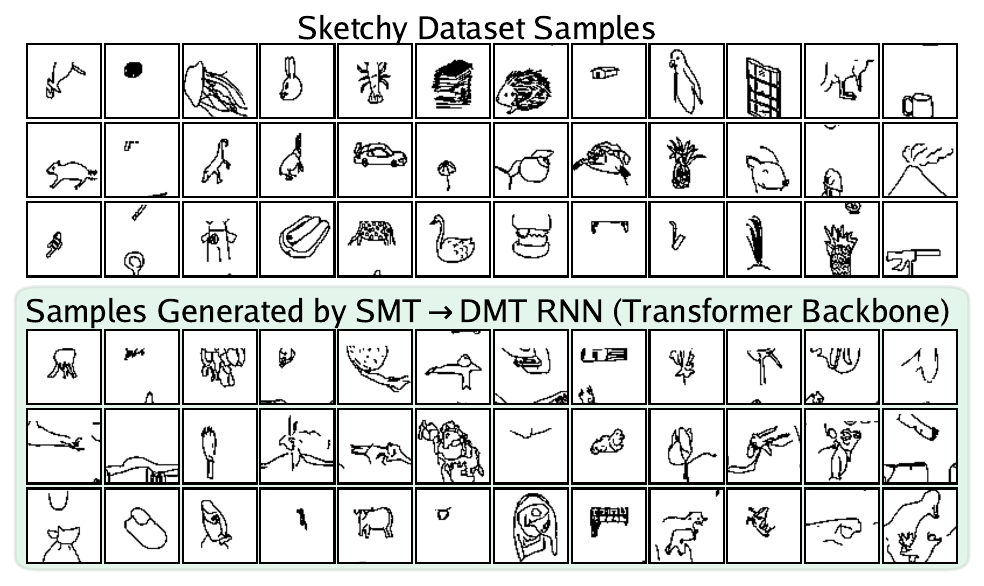}
    \captionof{figure}{%
      \textbf{Attneave's Sketchy Generation.}
      SMT$\rightarrow$DMT captures the stroke structure of human-drawn sketches through only pixel sequence modeling on sparse images.
      More samples are in Appendix Figure~\ref{fig:sketchy_generations_extra}.
    }
    \label{fig:sketchy_generations}
  \vspace{-30pt}
  \end{minipage}
\end{wrapfigure}

We now evaluate on Attneave's tasks.
Figure~\ref{fig:mnist_generations} shows the stark difference between MNIST samples generated by RNNs trained with BPTT and SMT$\rightarrow$DMT.
Figure~\ref{fig:sketchy_generations} shows images generated by an SMT$\rightarrow$DMT RNN trained on Sketchy.

Along with the synthetic experiments, these results confirm that SMT doesn't suffer from a recency bias like BPTT, allowing it to properly attribute credit across long sequences.

\vspace{-5pt}
\subsection{Sequential Compute and Data}
\label{sec:seqflops_experiments}
\vspace{-10pt}

We now evaluate BPTT and SMT across real domains and various RNN architectures.
Each method is allowed $N$ optimization steps on token batches of shape $B\times T$ (number of sequences $\times$ sequence length).
We sweep $N$, $B$, and $T$ for each method to profile how much sequential compute and data each method uses to achieve a target performance.
Sequential compute, measured in sequential FLOPs, is a metric proportional to the amount of inherently serial steps required to do the computation ($\sim$ time it would take on an infinitely parallel computer).
Sequential compute is a useful quantity because modern hardware is highly parallel, making it the primary constraint in large-scale model training~\citep{hooker2021hardware}.
Data is measured by the number of tokens processed by the model during training.
We elaborate on how sequential FLOPs and data is calculated for each method in Appendix~\ref{sec:appendix_definitions}.

Figure~\ref{fig:seqflops} shows the results.
In \textbf{sequential compute}, the SMT Encoder and SMT$\rightarrow$DMT RNN are significantly more efficient than BPTT across both datasets and most RNN backbones, which is expected due to its parallelizability.

In \textbf{data}, the SMT Encoder and SMT$\rightarrow$DMT RNN has approximately the same data efficiency as BPTT on TinyStories, a natural language dataset.
However on MNIST, the SMT Encoder and SMT$\rightarrow$DMT RNN shows significantly better data efficiency.
This result is explained by the short vs long range dependency requirements of natural language~\citep{fang2025wrongperplexitylongcontextlanguage} vs pixel sequence modeling~\citep{tay2020longrangearenabenchmark}.

SMT$\rightarrow$DMT is general enough to train a variety of RNN architectures, including ones with GRU/LSTM gated updates. 
However, Appendix~\ref{sec:additional_experiments} shows that it is unable to train \textit{vanilla stacked} GRU/LSTMs due to their hierarchical memory state structure, which is not fit for the encoder Transformer.

\begin{figure}[t]
    \centering
    \includegraphics[width=1.0\linewidth]{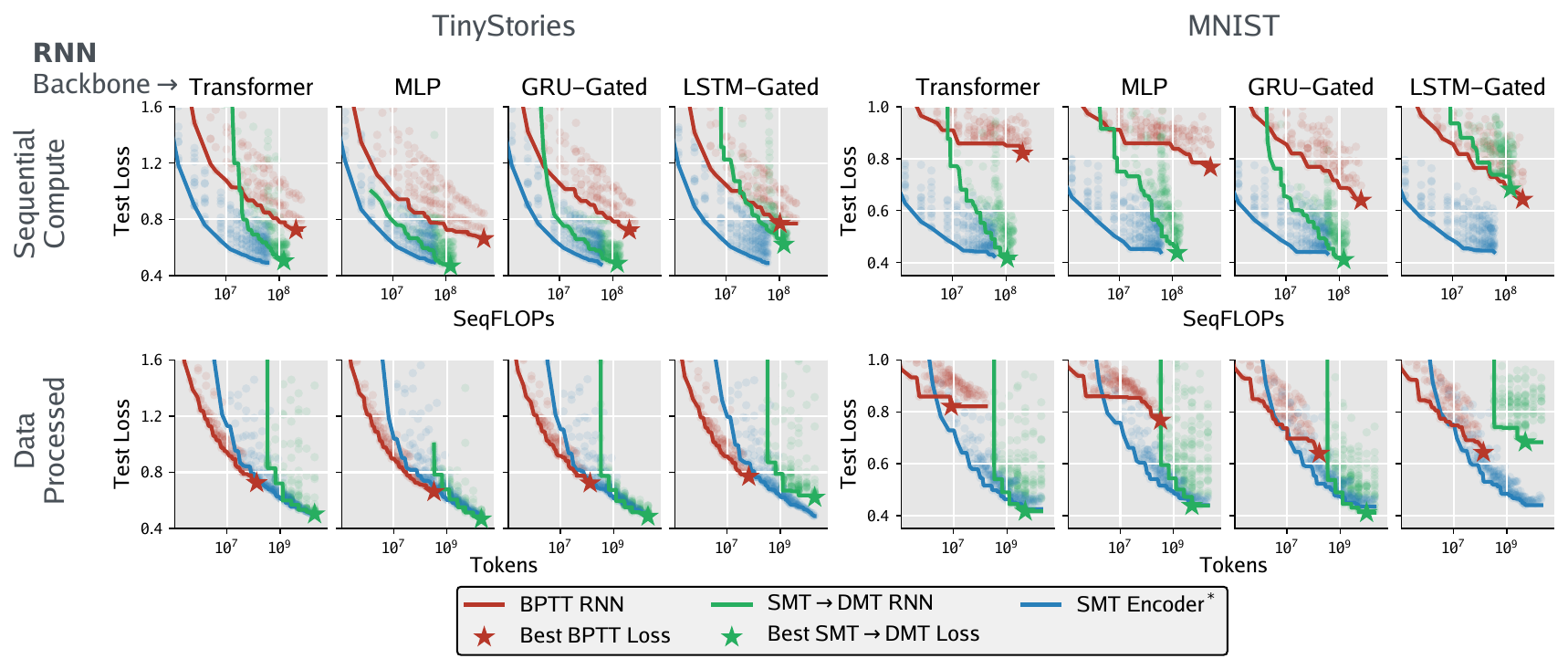}
    \caption{
    \textbf{Sequential Compute and Data Efficiency.}
    We sweep training hyperparameters for BPTT, SMT, and SMT$\rightarrow$DMT and plot the resulting runs' performance along sequential compute (SeqFLOPs) used and data processed (Tokens), across different RNN architectures and datasets.
    Runs are capped at one day on an H200 GPU.
    $^*$ signifies that the SMT Encoder is the teacher Transformer (\textit{not} an RNN) and is shown only as a reference.
    In sequential compute, SMT and SMT$\rightarrow$DMT are generally more efficient than BPTT across datasets and architectures.
    In data efficiency, SMT and SMT$\rightarrow$DMT are around the same as BPTT on TinyStories, but more efficient on MNIST, likely due to the long-range credit assignment required for MNIST pixel sequence modeling.
    Figure~\ref{fig:seqflops_wallclock} provides plots against total FLOPs and wallclock time.
    }
    \label{fig:seqflops}
    \vspace{-15pt}
\end{figure}

\vspace{-8pt}
\subsection{Scaling Laws}
\label{sec:scalinglaws_experiments}
\vspace{-8pt}

\begin{figure}[b]
    \centering
    \begin{minipage}{0.43\textwidth}
        \centering
        \includegraphics[width=\linewidth]{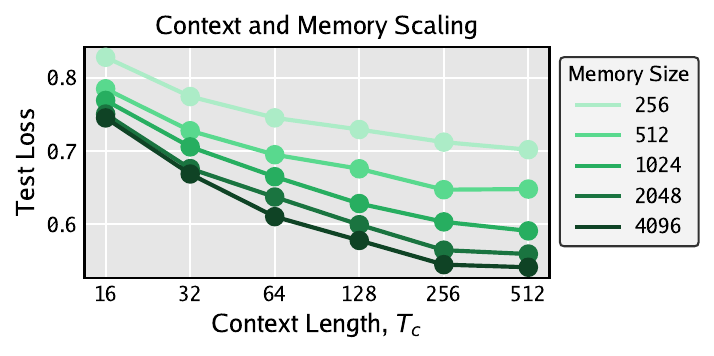}
        \caption{
        \textbf{Scaling Context and Memory.}
        SMT$\rightarrow$DMT shows smooth performance improvements as you increase the context length and the memory size in TinyStories.
        }
        \label{fig:memory_Tc_scaling}
    \end{minipage}
    \hfill
    \begin{minipage}{0.53\textwidth}
        \centering
        \includegraphics[width=\linewidth]{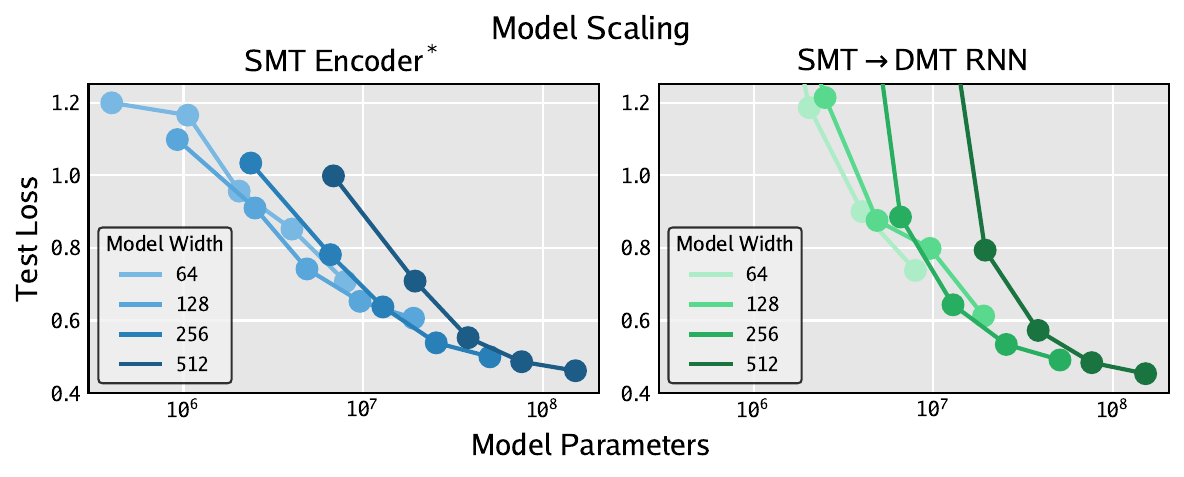}
        \caption{
        \textbf{Scaling Model Size.}
        Sweeping the width and depth of the RNN and teacher shows smooth performance improvements in TinyStories.
        The RNN imitates the teacher performance better at larger scale.
        }
        \label{fig:model_scaling}
    \end{minipage}
    \vspace{-10pt}
\end{figure}

We evaluate the scaling behavior of SMT$\rightarrow$DMT along three axes: context length, memory state size, and model parameter count.
For the first two, we logarithmically sweep $T_c$ and the number of memory tokens in the Transformer-based RNN (Figure~\ref{fig:architecture}).
For model scaling, we vary the width and depth of the RNN, encoder, and decoder.
We use the TinyStories domain for these experiments.

Figure~\ref{fig:memory_Tc_scaling} shows that SMT$\rightarrow$DMT exhibits smooth, predictable performance improvements with larger context length and bigger memory state size.
Together with the previous experiments, these results reaffirm that SMT effectively leverages long contexts and large memory states.
Figure~\ref{fig:model_scaling} presents the parameter scaling results.
The SMT encoder follows a standard power-law-like scaling trend.
The SMT$\rightarrow$DMT RNN also improves smoothly with scale, albeit with a differently shaped scaling curve.
Interestingly, the RNN appears to more closely match the encoder’s performance at larger scales.

\newpage

\subsection{Compression as a Scaling Axis}
\label{sec:compression_experiments}
\vspace{-5pt}

\begin{wrapfigure}{r}{0.45\textwidth}
    \centering
    \vspace{-35pt}
    \includegraphics[width=1.0\linewidth]{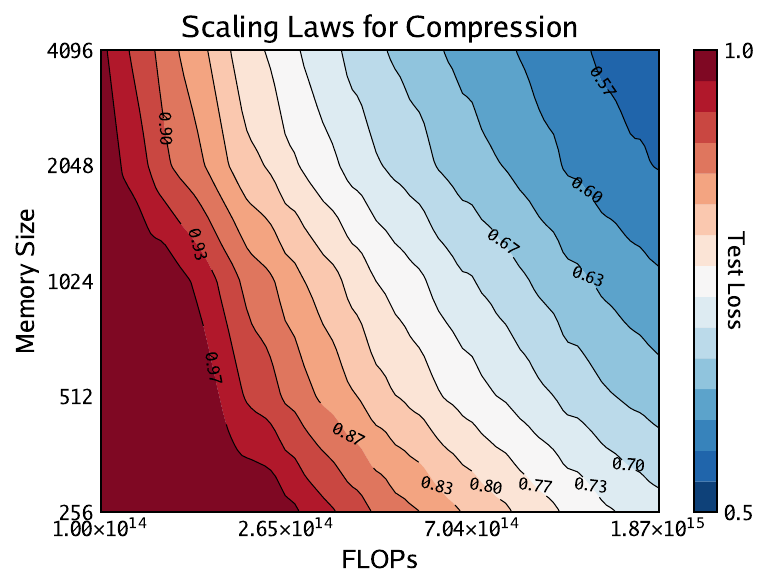}
    \vspace{-15pt}
    \caption{
    \label{fig:compression_scaling}
    \textbf{Scaling Laws for Compression.}
    We plot iso-loss contours for SMT-trained encoder models across a range of memory state sizes and training compute budgets.
    For a fixed target performance, SMT can achieve higher compression (smaller memory size) using additional compute.
    This result suggests a new property to scale when given more training compute: memory state compression.
    }
    \vspace{-35pt}
\end{wrapfigure}

Neural scaling laws predict the relationship between a resource (e.g. compute, data) and a desired property (e.g. validation loss, benchmark accuracy)~\citep{kaplan2020scaling,hoffmann2022training}.
Can the desired property instead be \textit{compression}~\citep{hutter2005universal}?
For RNNs, compression can be interpreted as achieving the same performance with a smaller memory state size.
Thus, to answer this question, we train a set of SMT models on TinyStories across a sweep of memory state sizes and training compute budgets.

Figure~\ref{fig:compression_scaling} shows the scaling curve, confirming that SMT can achieve more compression when allocated more compute.
Since compression is often speculated as being a core property of intelligent systems~\citep{solomonoff1964formal,kolmogorov1968three}, scaling along this compression axis may be a desired direction forward for future sequence models.
Notably, Transformers perform no compression of the past~\citep{gu2025tradeoffs}, which may explain their training efficiency.

\vspace{-5pt}
\subsection{Ablations}
\label{sec:ablations_experiments}
\vspace{-5pt}

\paragraph{Predictive State and Detached RNN}
\label{sec:detach_rnn_experiments}

The impact of the predictive state objective (Equation~\ref{eq:loss_dec}) is evaluated by sweeping the future length $T_f$, while keeping $T_c$ large enough to see the whole sequence.
The impact of the dynamics objective (Equation~\ref{eq:loss_dyn}) on memory representation is tested by detaching the model computation graph with stop grads at two locations such that the gradients from $\mathcal{L}_t^\text{dyn}$ flow to the RNN, but \textit{not} the encoder (\textbf{detached}); the non-detached SMT baseline is referred to as \textbf{joint}.
This ablation isolates the contribution of explicitly training $m_t$ to be a Markovian representation.

\begin{figure}[b]
    \vspace{-15pt}
    \centering
    \begin{minipage}{0.40\linewidth}
        \centering
        \includegraphics[width=\linewidth]{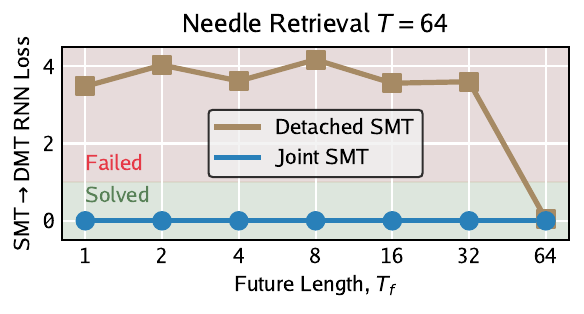}
        \caption{
        \textbf{Joint SMT Ablation.}
        Here, the task requires credit assignment across $T$ timesteps.
        When the RNN is detached during SMT, $T_f$ must be large enough to capture the task signal ($T_f=T$).
        With joint training, SMT solves the task even when $T_f$ is small.
        }
        \label{fig:needle_tf_sweep}
    \end{minipage}
    \hfill
    \begin{minipage}{0.56\linewidth}
        \centering
        \includegraphics[width=\linewidth]{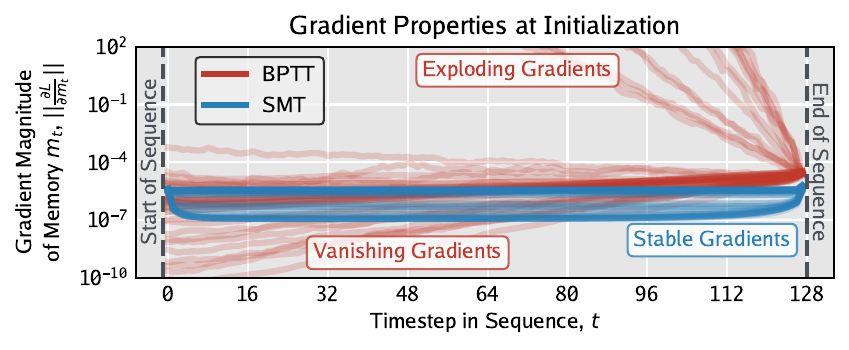}
        \vspace{-18pt}
        \caption{
        \textbf{Gradient Properties of BPTT and SMT.}
        In the needle retrieval task, the loss is applied at the last timestep.
        BPTT propagates gradients backward through all timesteps, risking vanishing/exploding gradients for each $m_t$, depending on the weight initialization.
        SMT is non-recurrent and has a $\mathcal O(1)$ credit path length, making its gradients agnostic to initialization and time-horizon.
        }
        \label{fig:gradient_vanishing}
    \end{minipage}
    \vspace{-12pt}
\end{figure}

Figure~\ref{fig:needle_tf_sweep} shows the results on the needle retrieval task.
To solve the task, and thus have proper credit assignment, SMT requires either large enough $T_f$ \textit{or} joint training.
When $T_f$ is large enough, there is an $\mathcal O(1)$ credit path length between the needle and the answer at all timesteps.
Interestingly, when $T_f$ is small, there exists no credit path to learn early timestep memories, yet joint training still learns effectively, even when $T_f=1$.
Credit must be propagating through the RNN dynamics from $m_{T}$ to $m_{T-1}$, and so on, to $m_0$.
But because the RNN is never unrolled, there is no computation graph for credit to propagate directly.
The only explanation is that \textbf{credit is being amortized into gradient optimization steps}.
Each optimization step sends information from $m_t$ to $m_{t-1}$ through $f_\theta$; $T$ such gradients steps sends information $T$ steps back in the sequence.
This implies that solving $T$ sequence length credit assignment task when $T_f=1$, requires at least $T$ \textit{gradient optimization steps}.
This credit amortization phenomenon is reminiscent of value bootstrapping in RL~\citep{sutton1998reinforcement}.

\vspace{-5pt}
\paragraph{$\lambda$ Coefficients}
The values of $\lambda_\text{dyn}$ and $\lambda_\text{unif}$ are swept here to check their effects.
Figure~\ref{fig:lambda_ablation} shows the results.
The best RNNs require $\lambda_\text{dyn}= 0.1$, and $\lambda_\text{unif}=0.001$.
When $\lambda_\text{unif}=0$, although the RNN performance is preserved, the memory space is collapsed, as indicated by $\mathcal L^\text{unif}$.

\begin{figure}
    \centering
    \includegraphics[width=1.0\linewidth]{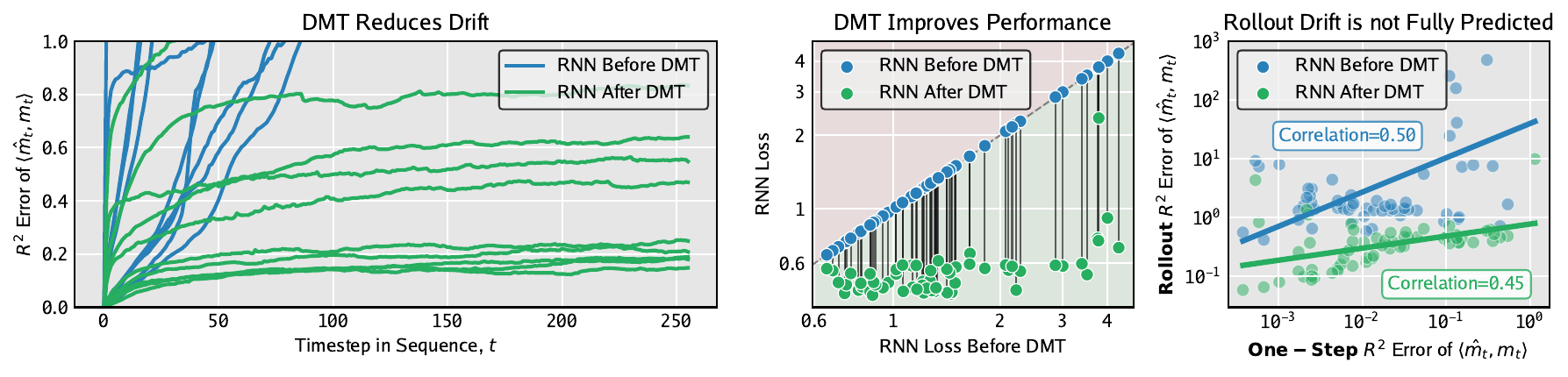}
    \caption{
    \textbf{Impact of DMT} across many runs with different SMT $\lambda_\text{dec}$ and $\lambda_\text{dyn}$ hyperparameters.
    \textbf{Left:}
    Applying DMT reduces the drift of the RNN rollout (measured with $1-R^2$ of RNN memory prediction $\hat m_t$ of encoder ground truth $m_t$).
    \textbf{Middle:}
    DMT significantly improves RNN performance across settings.
    \textbf{Right:}
    The one-step drift of the RNN only partially correlates with the rollout drift.
    \vspace{-10pt}
    }
    \label{fig:dmt_analysis}
\end{figure}

\vspace{-5pt}
\paragraph{Drift and DMT}
As described in Section~\ref{sec:dmt}, RNN suffers from drift post-SMT.
Figure~\ref{fig:dmt_analysis} shows an analysis of drift and DMT's mitigation of it.
From a dynamical systems perspective, DMT seems to discover RNNs which have an initially higher drift, but which \textit{plateau} at a much lower equilibrium drift.
Interestingly, this equilibrium drift value is not fully predicted by the one-step drift, inviting future investigations into predicting and mitigating rollout drift in one-step during SMT.

\vspace{-5pt}
\subsection{Analysis}
\label{sec:analysis_experiments}
\vspace{-5pt}

\paragraph{Gradient Properties of BPTT and SMT}
\label{sec:vanishing_grads}
The fundamental difference between BPTT and SMT in long-range credit assignment is dictated by their gradients.
Figure~\ref{fig:gradient_vanishing} shows the gradient magnitude of $m_t$, $\|\frac{\partial L}{\partial m_t}\|$, at different $t$ for both methods with different model weight initializations on the needle retrieval task.
In BPTT, gradients vanish or explode over time, due to BPTT's gradient propagation through recurrent modules.
In SMT, gradient magnitudes are independent of $t$, because the credit path length between tokens is independent of the sequence length. %
This result explains why SMT does not suffer a recency bias and is able to stably perform long-horizon credit assignment.

\begin{wrapfigure}{r}{0.45\textwidth}
    \centering
    \vspace{-15pt}
    \includegraphics[width=1.0\linewidth]{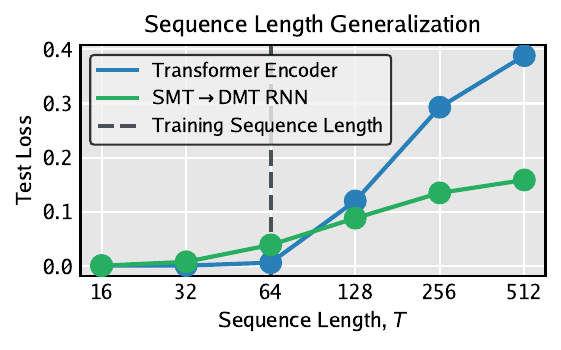}
    \caption{
    \label{fig:seqlen_generalization}
    \textbf{Sequence Length Generalization.}
    An SMT$\rightarrow$DMT trained RNN generalizes better than its Transformer teacher when evaluated on sequence lengths longer than training.
    The task is synthetic state tracking.
    }
    \vspace{-10pt}
\end{wrapfigure}

\vspace{-8pt}
\paragraph{Benefit of RNNs over Transformers}
SMT trains an RNN to mimic a Transformer encoder model, raising the question of why an RNN is needed at all, given the Transformer.
RNNs are qualitatively more efficient than Transformers at inference, requiring $O(1)$ rather than $O(T)$ memory and compute per generated token (Table~\ref{tab:resources_bigo}).
RNNs also constitute a more expressive class of models~\citep{merrill2026linear,liu2025serial}.

Here, we compare an SMT$\rightarrow$DMT RNN against a Transformer on the synthetic stack state tracking task.
For a fair comparison, we use the SMT encoder as the Transformer baseline, since it imposes the same memory-information bottleneck as the RNN.

Figure~\ref{fig:seqlen_generalization} shows the Transformer outperforms the RNN on training sequence lengths, but significantly underperforms the RNN on sequence lengths longer than training.
Prior work on length generalization reports similar findings~\citep{press2021train}.
This result reflects the distinct inductive biases of the architectures: Transformers behave like growing lookup tables in context, while RNNs update finite states~\citep{gu2025tradeoffs}.
The latter is a better inductive bias for generalization.

\begin{figure}
    \vspace{-5pt}
    \centering
    \begin{minipage}{0.65\linewidth}
        \centering
        \includegraphics[width=\linewidth]{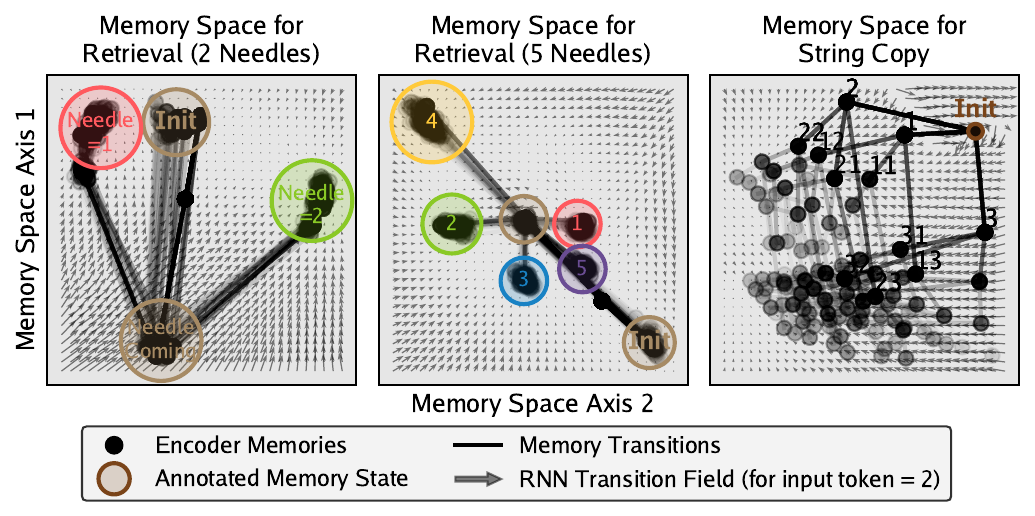}
    \end{minipage}
    \hfill
    \begin{minipage}{0.32\linewidth}
        \caption{
        \textbf{Memory Space Visualization.}
        The encoder learns different memory geometries for different tasks.
        In Retrieval, the encoder collapses many sequence states into a few memory states, creating finite-state machine like behavior.
        In String Copy, the encoder constructs a tree-like memory geometry to compress all possible sequences.
        Some geometries induce more complex RNN transition fields.
        }
        \label{fig:memory_space}
    \end{minipage}
\vspace{-20pt}
\end{figure}

\vspace{-8pt}
\paragraph{Memory Space}
To better understand what SMT is learning, we train smaller SMT models that have a 2D memory state and directly visualize their memory space across three synthetic tasks in Figure~\ref{fig:memory_space}.
In the retrieval tasks, SMT learns to collapse many sequence states into only a few effective memory states: an initial state, a state indicating the next token is the needle, and states corresponding to the needle value.
Then, the RNN learns finite-state machine behavior to transition between these states.
In contrast, string copying requires lossless sequence compression and thus SMT cannot alias distinct memory states together.
It learns to create a tree-like memory geometry to store all possible sequences, matching the tree structure of all possible strings.
Figure~\ref{fig:mnist_memory_pca} and Figure~\ref{fig:mnist_memory_tsne} show memory visualizations for models trained on MNIST.
These results indicate SMT memories form effective temporal \textit{abstractions} of the past depending on what the future requires.

\vspace{-5pt}
\paragraph{Additional Experiments}
Appendix~\ref{sec:additional_experiments} provides more results and analysis of SMT, including a comparison with a ParaRNN (DEER)~\citep{danieli2025pararnnunlockingparalleltraining} baseline.

\section{Related Works}
\label{sec:related_works}
\vspace{-5pt}

\paragraph{Relation with NextLat}
Highly related and concurrent to our work,~\citet{teoh2025nextlatentpredictiontransformerslearn} introduced Next-Latent Prediction (NextLat), which co-trains an RNN with memory state supervision from a Transformer.
With a particular setting of hyperparameters, NextLat closely resembles DMT (and SMT when the rollout length is 1), though still utilizing BPTT on multi-step rollouts.
However, the central focus of NextLat is to regularize the Transformer to learn compact world models, rather than providing a method to train RNNs.
Consequently, their analysis focuses primarily on the Transformer.
In contrast, our work focuses exclusively on training the RNN in a one-step manner.

\paragraph{Recurrent Neural Networks (RNNs)}
RNNs were studied extensively early in AI because their recurrence mechanism resembles biological brains~\citep{mcculloch1943logical} and can be applied to any sequential task~\citep{elman1990finding}.
Many different algorithms were proposed for learning, including random guessing~\citep{schmidhuber2001evaluating}, evolutionary algorithms~\citep{miller1989designing,angeline1994evolutionary,stanley2002evolving,salimans2017evolution,sarkar2025evolution}, hebbian learning~\citep{hebb1949organization,hopfield1982neural,miconi2018differentiableplasticitytrainingplastic,najarro2022metalearninghebbianplasticityrandom}, real-time recurrent learning~\citep{williams1989learning}, and other algorithms~\citep{ollivier2015training,bengio2015scheduledsamplingsequenceprediction,lamb2016professor,bai2019deep,kag2021training}.
BPTT is the only widely adopted algorithm~\citep{werbos1990backpropagation}.

However, it has repeatedly been shown that BPTT produces unstable gradients that vanish, explode, or exhibit high variance~\citep{bengio1994learning,hochreiter2001gradient,pascanu2013difficulty,bai2018empirical}.
Several directions address this issue.
One direction focuses on architectural modifications, including residual connections~\citep{srivastava2015highway,he2016deep} and gating mechanisms~\citep{chung2014empirical}, culminating in the development of the LSTM~\citep{hochreiter1997long} and GRU~\citep{cho2014properties}.
A parallel direction addressed gradient instability through orthogonal weight parameterizations to prevent exponential growth or decay across time~\citep{saxe2013exact,arjovsky2016unitary,wisdom2016full,mhammedi2017efficient,vorontsov2017orthogonality,helfrich2018orthogonal}.
Others explored external memory~\citep{peshkin2001learning,graves2014neural,hansen2017longtimescalecreditassignment}, hierarchical modeling~\citep{hihi1995hierarchical,chung2016hierarchical,wang2025hierarchicalreasoningmodel,jolicoeurmartineau2025morerecursivereasoningtiny}, other unique directions~\citep{jaeger2001echo,lukovsevivcius2009reservoir,miller2018stable}.

Recently, there has been renewed interest in RNNs in the form of linear state space models~\citep{gu2021efficiently,smith2022simplified,gu2023mamba}, linear attention models~\citep{katharopoulos2020transformers,sun2023retentive, dao2024transformers,yang2025parallelizinglineartransformersdelta,yang2025gateddeltanetworksimproving}, and even nonlinear RNN models~\citep{beck2024xlstm,bulatov2022recurrent,mishra2026m,oncescu2026recurrent}.
Recurrent computation more generally has been reappearing across paradigms including in diffusion~\citep{ho2020denoising}, looped Transformers~\citep{giannou2023looped}, and reasoning~\citep{hao2025traininglargelanguagemodels,geiping2025scaling}.

\vspace{-8pt}
\paragraph{Time-Parallel Training}

Transformers revolutionized sequence modeling~\citep{brown2020language} largely because they have time-parallel training~\citep{vaswani2017attention} (unlike prior attention methods~\citep{bahdanau2016neuralmachinetranslationjointly}), which is crucial for leveraging modern hardware~\citep{hillis1986data,hooker2021hardware} to scale performance~\citep{kaplan2020scaling,hoffmann2022training}.
Linear RNNs gained popularity~\citep{katharopoulos2020transformers,dao2024transformers,yang2025parallelizinglineartransformersdelta,feng2024were} after it was realized they can be parallelized with the associative scan algorithm~\citep{blelloch1990prefix,martin2018parallelizinglinearrecurrentneural,wang2020bppsascalingbackpropagationparallel}.

A recent line of work attempts to parallelize nonlinear RNNs as well~\citep{lim2024parallelizingnonlinearsequentialmodels,danieli2023deeppcr}.
Rather than computing $[m_0,\dots,m_T]$ with $m_{t+1}=f_\theta(m_t)$, they formulate the forward pass as an iterative optimization procedure.
Starting with an initial guess $[m_0^0,\dots,m_T^0]$, they construct a system of $T$ equations, $\{m_{t+1} - f_\theta(m_t)= 0\}_{t=0}^T$, and solve this system with Newton's method~\citep{ortega2000iterative}.
Many works have further built on this approach~\citep{danieli2025pararnnunlockingparalleltraining,gonzalez2024towards}.
Although appealing, this approach approximates BPTT and hence will suffer from its $\mathcal{O}(T)$ credit path length and corresponding gradient instability, along with the added convergence worries of Newton's method~\citep{gonzalez2025predictability}.
In contrast, SMT uses an encoder to train $m_t$ to be a predictive state while satisfying $m_{t+1}\approx f_\theta(m_t)$ and providing an $\mathcal{O}(1)$ credit path length.

\vspace{-8pt}
\paragraph{Computation Complexity Class of Models}
A model's architecture determines the problems it can theoretically solve~\citep{hornik1989multilayer}.
Some tasks are inherently sequential and cannot be efficiently parallelized~\citep{amdahl1967validity}; the \textbf{circuit depth} of a task is the minimum number of sequential steps required to solve it on an infinitely parallel computer~\citep{ruzzo1981uniform,cook1985taxonomy}.
Every neural network has a corresponding \textbf{sequential depth}---the longest nonlinear computation path from input to output---which bounds the class of problems it can solve~\citep{merrill2024expressivepowertransformerschain}.
Models with constant or logarithmic sequential depth per layer, such as Transformers and linear RNNs, are provably limited to tasks with equivalently low circuit depth~\citep{merrill2023parallelism,merrill2024expressivepowertransformerschain,yau2026sequentialparalleldualityprefixscannable}.
While such models succeed on tasks amenable to parallelization (e.g. parity tracking via associative scan~\citep{liu2022transformers,li2025howlanguagemodelstrack}), they systematically fail on tasks requiring deep sequential computation (e.g. tracking a chess board~\citep{merrill2024illusion}).
Interestingly, the aspect that makes models parallelizable, limits their performance on harder problems~\citep{merrill2023parallelism}.
Nonlinear RNNs are one of the few classes of models where its sequential depth grows with the input sequence length~\citep{merrill2026linear}.
Although these constraints were seen as theoretical, there is growing evidence they affect models in practice as well~\citep{liu2025serial}.

In SMT, we train a nonlinear RNN (which is fully expressive), using a time-parallel teacher Transformer (which has limits).
We note this limitation but argue that SMT is a pretraining algorithm, which should be used with a lightweight post-training algorithm to solve downstream tasks~\citep{gan2026neural}.

\vspace{-5pt}
\paragraph{Predictive State Representations (PSRs)}
A PSR is a way of modeling a partially observed dynamical system by representing its state only in terms of predictions about future observations~\citep{littman2001predictive,singh2012predictive}, a representation that is sufficient for optimal decision making~\citep{singh2003learning}.
Early works interpreted PSRs as a literal vector of probabilities of future events, but have since been generalized~\citep{downey2017practical}.
Belief states are a similar concept, which also defines a sufficient statistic of the past~\citep{kaelbling1998planning,hu2024learning}.

PSRs have been previously incorporated into RNNs~\citep{downey2017predictivestaterecurrentneural,hefny2018recurrentpredictivestatepolicy}.
\citet{venkatraman2017predictivestatedecodersencodingfuture} introduce an auxiliary objective for RNNs that trains hidden states to predict statistics of future observations using a decoder.
However, these works still unroll the RNN and use BPTT, and thus are not time-parallelizable and have a $\mathcal{O}(T)$ credit path.

\paragraph{Other Related Work}
Our work is related to the literature on cross-architecture teacher-student distillation~\citep{kasai2021finetuningpretrainedtransformersrnns,wang2025mamballamadistillingaccelerating,hauzenberger2026effectivedistillationhybridxlstm,chen2026hybridlinearattentionright,moudgil2026attention}, but these works do not address the challenges of training nonlinear RNNs.
The Recurrent Transformer is an RNN architecture that attends to all past hidden states, creating an $\mathcal{O}(1)$ gradient path that stabilizes credit assignment~\citep{oncescu2026recurrent}.
However, because it retains all past hidden states, its memory grows unboundedly during inference---making it more akin to a Transformer than a fixed-memory RNN.
Crucially, training still requires sequential unrolling and BPTT.
SMT, by contrast, replaces BPTT and supports arbitrary fixed-memory RNN architectures and enables time-parallel training by never unrolling the RNN.
Other works similarly combine Transformers with recurrent processing but also train with sequential unrolling and BPTT~\citep{fan2020addressing,bulatov2022recurrent,wu2022memformer}.
A new line of work uses principles from diffusion models to train blocks of a feed-forward network in parallel, avoiding global backpropagation~\citep{li2025noprop,shing2025diffusionblocks}.

\vspace{-10pt}
\section{Discussion}
\label{sec:discussion}
\vspace{-10pt}

In SMT, the teacher model is time-parallel, and is thus constrained in expressivity~\citep{merrill2023parallelism}, implying that SMT-trained RNNs may suffer the same problem.
Therefore, BPTT finetuning might be required to achieve expressivity beyond the teacher.
Additionally, while SMT is useful for learning how to encode sequences, it is not necessarily to be used for learning reasoning since intermediate steps are not supervised.
The same limitation applies to Transformers yet post-training allows them to effectively solve longer-horizon tasks than the training horizon; the same might be true for SMT-trained RNNs.

The current SMT variant computes and trains only a single $m_t$ within a sequence.
We found that training on all memories $[m_0, \dots, m_T]$ offered no improvement in our settings, but this may not hold at larger scales.
After SMT, the RNN experiences drift away from the teacher memory trajectory.
DMT provides one solution but is not time-parallel; however, it may be parallelized via DEER~\citep{lim2024parallelizingnonlinearsequentialmodels}.

RNNs have the promise of solving problems that extend over unbounded horizons, such as the entire lifetime of an agent.
However, training methods for RNN have been hindered by the inability of BPTT to assign credit effectively over such a long horizon.
Our method circumvents the credit assignment issue with an $\mathcal O(1)$ connection path.
In the regimes we studied, this effectively allows for learning memories that are only useful many steps later, an ability that is crucial for lifelong learning.

\newpage

\vspace{-5pt}
\begin{ack}
\vspace{-5pt}
This work was supported by an NSF GRFP Fellowship to A.K., a Packard Fellowship and Sloan Research Fellowship to P.I., and ONR MURI grant N00014-22-1-2740.
This work was also supported under project ID 43 as part of the Swiss AI Initiative, through a grant from the ETH Domain and computational resources provided by the Swiss National Supercomputing Centre (CSCS) under the Alps infrastructure.
\\
We thank Alyosha Efros for suggesting the Attneave framing for pixel sequence modeling and recommending the Sketchy dataset.
We thank Alexander Huth for initially motivating A.K. to work on memory many years ago.
We thank Assaf Ben-Kish for reviewing an earlier draft of this manuscript.
We thank Han Guo and Oliver Sieberling for technical advice on algorithmic complexity.
\end{ack}

\bibliographystyle{plainnat}
\bibliography{references}

\appendix

\newpage

\section{Definitions}
\label{sec:appendix_definitions}

\vspace{-5 pt}
\paragraph{Credit Assignment Path Length}
For any differentiable computational graph, backpropagation propagates gradients from the scalar loss backward through the graph to each leaf node (typically model weights).
We define the \textbf{credit assignment path length} as the maximum distance between any two nodes (e.g. tokens) in the computation graph.
Distance is measured as the number of intervening non-identity operations that modify gradients (e.g. matrix multiplications or nonlinearities).
The longer this path, the less effective backpropagation is for properly learning associations between distant nodes and assigning credit~\citep{bengio1994learning}.
Under this definition, BPTT has $\mathcal O(T)$ credit assignment path length, whereas Transformers and SMT have $\mathcal O(1)$ path length between any two tokens.

\vspace{-5 pt}
\paragraph{Sequential Computation (measured in SeqFLOPs)}
Sequential computation is the amount of serial (non-parallelizable) work required to complete a computation.
Some computations may require substantial total work but little sequential work (e.g. matrix multiplication).
As parallel hardware such as GPUs continues to improve, total work matters less than the amount of inherently sequential work required~\citep{hooker2021hardware}.

Sequential compute is measured by analyzing the computation graph required to execute an algorithm and computing the graph’s \textit{critical path}: the number of floating point operations that must be executed sequentially on an infinitely parallel computer.
We refer to this quantity as \textbf{sequential FLOPs} (SeqFLOPs).

For simplicity, we estimate SeqFLOPs by counting the number of sequential atomic deep learning operations (e.g., Linear, LayerNorm) executed over the course of the algorithm.
The true SeqFLOPs, measured in floating-point operations, is proportional to this estimate.

We compute SeqFLOPs for BPTT, SMT, and DMT.
SMT is fully parallelizable in time, incurring $\mathcal{O}(1)$ SeqFLOPs per optimization step, independent of the sequence length $T$.
In contrast, BPTT and DMT require unrolling the RNN, which increases SeqFLOPs to $\mathcal{O}(T)$ per optimization step.

\vspace{-5 pt}
\paragraph{Data Processed (measured in Tokens)}
Data Processed is defined as the total number of tokens processed during training, including repeated tokens in multi-epoch settings.

\section{Experiment Details}
\label{sec:appendix_experiment_details}

\subsection{Architectures}
\label{sec:appendix_experiment_details_archs}
\begin{figure}
    \centering
    \includegraphics[width=1.0\linewidth]{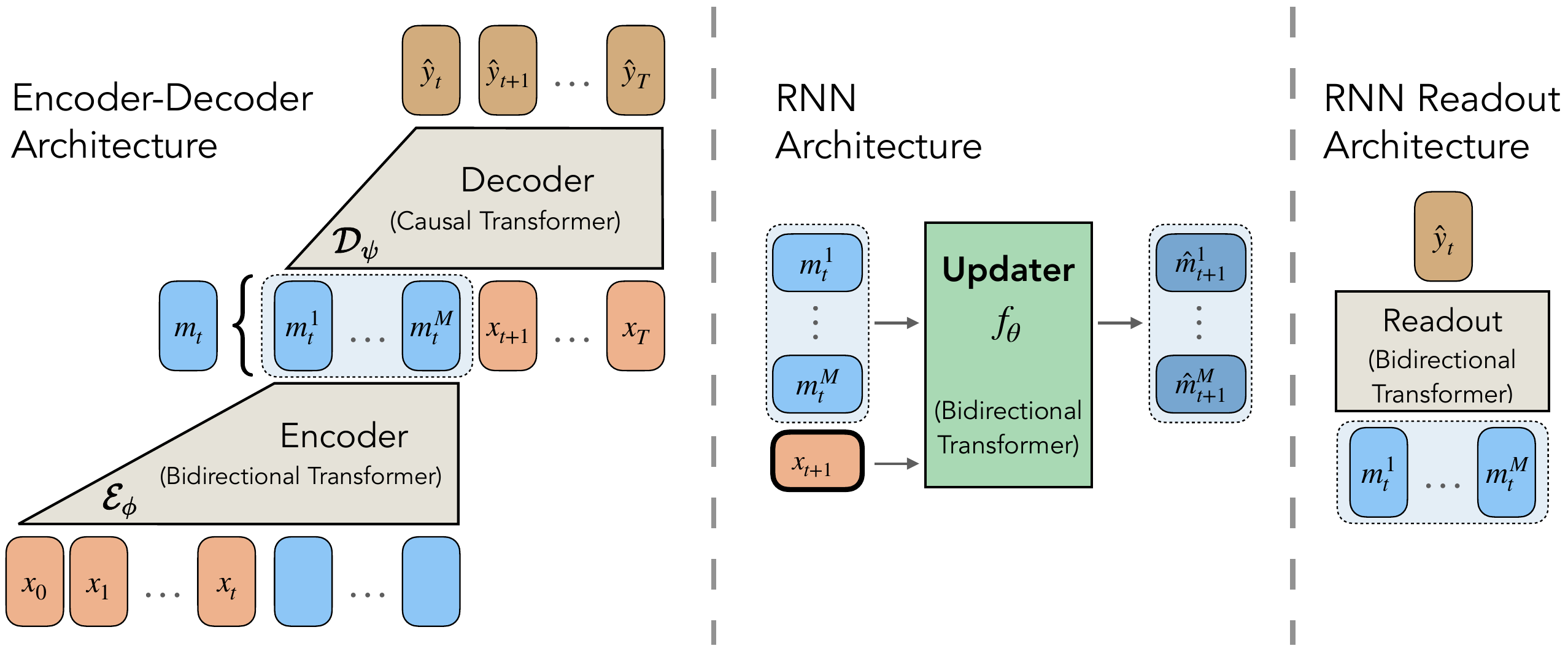}
    \caption{
    \textbf{Model Architecture for SMT.}
    \textbf{Left:}
    The encoder reads the input context tokens and a set of learned register tokens, and outputs the memory, $m_t$, which is a set of memory tokens.
    The decoder takes in this memory and the \textit{future input} tokens and predicts the \textit{future output} tokens, using a causal mask.
    This setup forces information from the context to be compressed into a memory that is useful for predicting the future outputs, given future inputs.
    \textbf{Middle:}
    Our RNN maps $(m_t, x_{t+1})$ to $m_{t+1}$ using a Transformer-backbone.
    Since the memory is a list of tokens and the input is a token, we simply use a full attention Transformer to transform the current memory into the next timestep's memory.
    \textbf{Right:}
    Readout is performed by a full attention Transformer over the memory tokens.
    }
    \label{fig:architecture}
\end{figure}

Our primary architecture used for most experiments is shown in Figure~\ref{fig:architecture}.

\vspace{-5 pt}
\paragraph{Encoder Architecture}
We use the same encoder model architecture across all experiments.
The model begins with an embedding layer to embed input tokens, and a list of learned memory token registers.
The input and register tokens are concatenated and processed by a stack of \textit{bidirectional} (full attention mask) Transformer blocks.
We use bidirectional model because the goal is to create a holistic representation of the entire input sequence.
The register tokens are then interpreted as memory tokens at the output.
Note that a single memory consists of a list of memory tokens, $m_t=[m_t^1,\dots,m_t^M]$.
Within a Transformer block, we use rotary position encodings~\citep{su2024roformer} and RMSNorm instead of LayerNorm.
We perform RMSNorm on the output memory tokens for stability.
For the GRU-Gated, LSTM-Gated, stacked GRU, and stacked LSTM RNN architectures, we perform tanh instead of RMSNorm on the output memory tokens, so the Encoder does not output memory tokens outside the range of what the RNN is capable of.
Figure~\ref{fig:architecture} shows the encoder architecture.

\vspace{-5 pt}
\paragraph{Decoder Architecture}
We use the same decoder model architecture across all experiments.
The decoder has an embedding layer to embed input tokens, which is weight shared with the encoder model.
The memory tokens from the encoder and the embedded future input tokens are concatenated, and then processed by a stack of \textit{causally masked} Transformer blocks.
We use a causal mask because the goal is to learn a generative model of the output sequence.
Within each Transformer block, we use rotary position encodings~\citep{su2024roformer} and RMSNorm instead of LayerNorm.
The output predictions are read out at token positions such that $\hat y_{t+k}$ is a function of only $m_t$ and $x_{t+1},\dots,x_{t+k}$.
Figure~\ref{fig:architecture} shows the decoder architecture.

\vspace{-5 pt}
\paragraph{Transformer-based RNN Architecture}
The Transformer-based RNN is our primary RNN architecture and is used for most experiments.
It begins with an embedding layer for the current timestep's input token.
The memory tokens are concatenated with the input token and then processed by a stack of \textit{bidirectional} (full attention mask) Transformer blocks to produce the output memory tokens.
We perform RMSNorm on the output memory tokens.

Note that although the architecture is a Transformer, this model acts as an RNN over the tokens in the sequence.
This RNN is similar in spirit to the Recurrent Memory Transformer~\citep{bulatov2022recurrent} and Looped Transformer~\citep{yang2024looped}.

\vspace{-5 pt}
\paragraph{RNN Readout Architecture}
We use the same readout model architecture for all RNNs.
The readout architecture takes in the memory tokens and processes them through a stack of \textit{bidirectional} (full attention mask) Transformer blocks.
Figure~\ref{fig:architecture} shows the readout architecture.

\vspace{-5 pt}
\paragraph{MLP-based RNN Architecture}
The MLP-based RNN flattens the list of memory tokens into a single vector, concatenates it with the input token embedding, and passes them through an MLP.
At the output, the model then unflattens them to be a list of memory tokens again.
We perform RMSNorm on the output memory tokens.

\vspace{-5 pt}
\paragraph{GRU-Gated RNN Architecture}
The GRU-Gated RNN uses the Transformer-based backbone, but gates the memory update according to the GRU gate equations.
Each gate variable is a nonlinear function of the previous memory and the input token, and is computed with Transformer backbones.

\vspace{-5 pt}
\paragraph{LSTM-Gated RNN Architecture}
The LSTM-Gated RNN uses the Transformer-based backbone, but gates the memory update according to the LSTM gate equations.
Each gate variable is a nonlinear function of the previous memory and the input token, and is computed with Transformer backbones.

\vspace{-5 pt}
\paragraph{Vanilla Stacked GRU RNN Architecture}
The stacked GRU RNN processes the $M$ memory tokens with a $M$ layer vertically stacked GRU.
Layer $l$ reads a single memory token, $m_t^l$, and outputs a single memory token for the next timestep, $m_t^{l+1}$.
We do \textit{not} do RMSNorm on the output memory tokens, since that would undermine the GRU's residual structure.

Note that vertically stacking GRU cells explicitly gives the memory $m_t=\{m_t^1,\dots,m_t^M\}$ a format of hierarchy.
$m_{t+1}^{l}$ is a function of $m_t^1,\dots,m_t^{l}$.
However, $m_{t+1}^{l}$ cannot access $m_t^{l+1},\dots, m_t^M$.
This hierarchical memory structure is not well suited for the SMT Encoder.

\vspace{-5 pt}
\paragraph{Vanilla Stacked LSTM RNN Architecture}
The stacked LSTM RNN processes the $M$ memory tokens with a $M/2$ layer vertically stacked LSTM.
Each layer processes two memory tokens, corresponding to the cell state and hidden state of the LSTM layer.
We do \textit{not} do RMSNorm on the output memory tokens, since that would undermine the LSTM's residual structure.

Same as the stacked GRU, vertically stacking LSTM cells explicitly gives the memory $m_t=\{m_t^1,\dots,m_t^M\}$ a format of hierarchy, which is not well suited for the SMT Encoder.

\subsection{Datasets}

\subsubsection{Synthetic Tasks}
\label{sec:synth_task_desc}

\vspace{-5 pt}
\paragraph{Retrieval to test \emph{Gradient Stability}}
The retrieval task requires the model to remember and reproduce the token immediately following a designated identifier (token $0$).
For example, given 
$\mathbf{x}=[3,4,0,2,1,3,1,0]$
the target is
$\mathbf{y}=[\emptyset,\emptyset,\emptyset,\emptyset,\emptyset,\emptyset,\emptyset,2]$, where $\emptyset$ denotes no prediction target.
With probability $p$, the label is corrupted to a random token.
By varying \textbf{sequence length} and \textbf{noise level}, this task probes the algorithm's capacity for stable gradient credit assignment.

\vspace{-5 pt}
\paragraph{String Copy to test \emph{Memory Capacity}}
The string copy task requires the model to reproduce a sequence in reverse order after a delimiter (token $0$).
For example, given
$\mathbf{x}=[3,4,1,2,0,0,0,0]$,
the target is
$\mathbf{y}=[\emptyset,\emptyset,\emptyset,\emptyset,2,1,4,3]$.
By varying the \textbf{sequence length} and the \textbf{memory state size}, this task measures the algorithm's ability to leverage the RNN's memory capacity for memorization.

\vspace{-5 pt}
\paragraph{Stack Operations to test \emph{State Tracking}}
The stack operations task requires the model to track the top element of a stack through a sequence of push and pop operations (denoted by token $0$).
For example, given
$\mathbf{x}=[1,0,2,3,0,1,0,0]$,
the target is
$\mathbf{y}=[\emptyset,1,\emptyset,\emptyset,3,\emptyset,1,2]$.
By varying the \textbf{sequence length} and \textbf{state complexity} (maximum stack depth), this task evaluates the algorithm's capacity for state tracking.

\vspace{-5 pt}
\paragraph{Keys and Values to test \emph{Associative Recall}}
The keys and values task requires the model to store and retrieve associations between keys and values, then recall the value corresponding to a queried key~\citep{willshaw1969non,hinton2014parallel}.
For example, given
$\mathbf{x}=[b,1,a,3,d,2,4,a]$,
the target is
$\mathbf{y}=[\emptyset,\emptyset,\emptyset,\emptyset,\emptyset,\emptyset,\emptyset,3]$.
By varying the \textbf{number of associations} and \textbf{association complexity} (string length of keys and values), this task evaluates the algorithm's capacity for associative recall.

\vspace{-5 pt}
\paragraph{Modular Arithmetic to test \emph{In-Context Learning}}
The modular arithmetic task requires the model to infer a latent linear rule from in-context examples and apply it to novel inputs~\citep{akyurek2022learning,kirsch2022general}.
For each sequence, parameters $a$ and $b$ are sampled, then the sequence is presented as
$\mathbf{x}=[x_0,y_0,x_1,y_1,x_2,y_2,x_3,y_3]$,
where $y_i=(a x_i + b) \mod V$, where $V$ is the vocabulary size.
Then, the target is
$\mathbf{y}=[y_0,\emptyset,y_1,\emptyset,y_2,\emptyset,y_3,\emptyset]$.
By varying the \textbf{difficulty} (range of values $a$, $b$ can take on) and the number of \textbf{in-context examples}, this task evaluates the algorithm's ability to induce in-context learning.

\subsubsection{Natural Tasks}

\paragraph{TinyStories}
TinyStories is a curated dataset of short stories generated by OpenAI’s GPT-4~\citep{eldan2023tinystories}.
We use ASCII character-level tokenization, yielding a vocabulary of 256 tokens.
Under this tokenization, the training and test sets contain 1.9B and 19.2M tokens, respectively.

\paragraph{MNIST}
MNIST is a classic image dataset consisting of handwritten digits~\citep{lecun1998mnist}.
Rather than performing classification or 2D image generation, we consider the problem of 1D pixel-sequence modeling.
The original 28$\times$28 images are flattened into sequences of length 784 using raster-scan ordering.
Each image is represented as a sequence of raw grayscale pixel intensities (0--255), yielding a vocabulary of 256 tokens.
The training and test sets contain 47M and 7.8M tokens, respectively.

\paragraph{Sketchy}
Sketchy is an image dataset of human-drawn sketches~\citep{sangkloy2016sketchy}.
Rather than performing classification or 2D image generation, we consider the problem of 1D pixel-sequence modeling.
The original images are resized to 64$\times$64 using Lanczos resampling, and the pixels are binarized.
Non-overlapping 2$\times$2 patches are tokenized, yielding a vocabulary of $2^{2\times2}=16$ tokens.
The resulting image is flattened in raster-scan order to form sequences of length 1024.
The training and test sets contain 69.5M and 7.7M tokens, respectively.

\subsection{Algorithms}
\label{sec:appendix_algorithm_details}
For all experiments, we use the AdamW optimizer with a weight decay of 0.01 and learning rates tuned separately for each algorithm.
For all methods, gradients are clipped to a maximum global norm of $1$.
Gradient clipping is expected to be particularly beneficial for BPTT.

After SMT, we transfer the decoder weights to the RNN readout module.
During DMT, this readout head is further finetuned to optimize the next-token prediction loss using RNN-generated memory states.
Importantly, this task loss updates only the readout head and not the RNN dynamics function, and therefore does not constitute temporal credit assignment for the RNN itself.
Instead, finetuning serves to adapt the readout head to imperfections in the memory states generated by the RNN.

\paragraph{Synthetic Experiments}
We set $T_c = T_f = T$.
To evaluate the expectation in $\mathcal{L}^\text{smt}$, we compute the loss terms at all timesteps $t\in[0, \dots, T]$.
For earlier timesteps, where the available past context is shorter than the required context length, we pad the sequence and modify the attention mask so that padding tokens are ignored.
The same procedure is applied to the future context.
In these synthetic experiments, the prediction loss is applied only at positions where target output tokens are defined (e.g.\ the answer token in the needle task).
We use batch sizes of 32 sequences during optimization, but this gets expanded to $32\times T$ input contexts to the encoder.

\paragraph{Other Experiments}
To evaluate the expectation in $\mathcal{L}^\text{smt}$, we compute the loss term at a single timestep sampled uniformly from $t \sim \mathcal U[0,\dots,T]$.
The dataset is represented as one long sequence, meaning padding is not required, as both the past and future contexts extend indefinitely.
We compute the $\mathcal{L}^\text{unif}$ over batches of memories from different sequences.

By default for SMT, we set $T_c = 256$, $T_f = 64$, $\lambda_\text{dec}=1.0$, $\lambda_\text{dyn}=0.1$, $\lambda_\text{unif}=0.001$ and train for $150000$ SGD iterations.
Unless otherwise specified, models use a hidden dimension of 256 with 16 memory tokens.
The encoder is 8 layers deep, while the decoder is 4 layers deep.
The RNN is also 8 layers deep, and its readout function is 4 layers deep.
We use a batch size of 128 sequences.

\newpage

\section{Additional Experiments}
\label{sec:additional_experiments}

\paragraph{Edge of Chaos}
Fundamental properties of dynamical systems (e.g. RNNs), including vanishing or exploding gradients, long-range memory, and the difficulty of imitating their dynamics, are closely tied to the \textit{predictability} of their dynamics~\citep{bengio1994learning,gonzalez2025predictability}. 
A standard measure of predictability is the Largest Lyapunov Exponent (LLE)~\citep{pikovsky2016lyapunov,gonzalez2025predictability}:
\begin{equation}
\text{LLE} = \lim_{T \to \infty} \frac{1}{T}
\log \left\| J_T J_{T-1} \cdots J_1 \right\|
\qquad \text{where} \qquad
J_t = \left.\frac{\partial f(s, x_t)}{\partial s}\right|_{s=s_{t-1}},
\end{equation}
for a system $s_t = f(s_{t-1}, x_t)$, assuming the limit exists (in practice, we use a finite-$T$ estimate).
The LLE measures the asymptotic rate at which small perturbations to the state grow or decay.
Negative LLEs indicate contracting dynamics, positive LLEs indicate exponential sensitivity to initial conditions (chaos), and LLEs near zero characterize systems potentially at the \textit{edge of chaos}~\citep{langton1990computation,bertschinger2004edge}.

Notably, the Jacobian product above is exactly the term through which BPTT propagates gradients, so negative LLEs imply vanishing gradients and positive LLEs imply exploding gradients~\citep{mikhaeil2022difficultylearningchaoticdynamics}.

We analyze the LLEs of the dynamical systems discovered by BPTT and SMT$\rightarrow$DMT.
We compute LLEs by propagating a tangent vector through the recurrent-state Jacobians using Jacobian-vector products, renormalizing at each timestep, and averaging the log expansion factors.
The SMT Encoder is not a recurrent system and so has no LLE in the conventional sense.
Instead, we compute a surrogate LLE by evaluating the SMT RNN's Jacobians along the state sequence produced by the Encoder rather than along the RNN's own trajectory.
Since Encoder states need not form a trajectory of the RNN, this quantity measures the average local expansion of the SMT RNN's dynamics in the region of state space the Encoder occupies.

\begin{figure}
    \centering
    \includegraphics[width=1.0\linewidth]{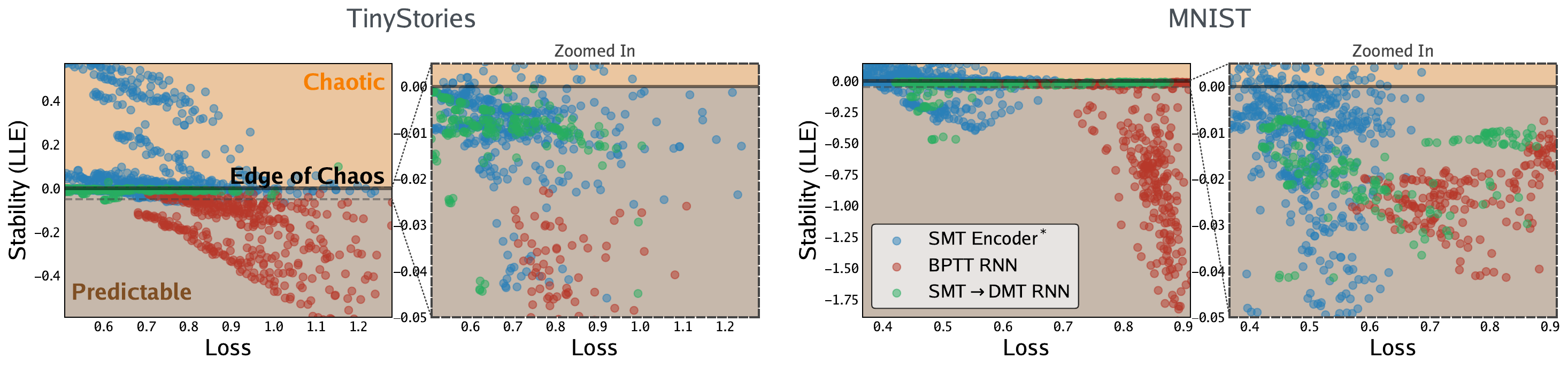}
    \caption{
    \textbf{SMT discovers RNNs at the Edge of Chaos.}
    We measure the Largest Lyapunov Exponent (LLE) of the dynamical system associated with the model from each method~\citep{gonzalez2025predictability}.
    BPTT often finds RNNs which are predictable (i.e. those with vanishing gradients).
    The SMT Encoder often seems to be at the edge of chaos or fully chaotic.
    Interestingly, SMT$\rightarrow$DMT consistently discovers RNNs which are at the edge of chaos.
    The SMT Encoder's LLE is estimated using the SMT RNN at the encoder's memory states.
    }
    \label{fig:lle}
\end{figure}

Figure~\ref{fig:lle} shows that BPTT and SMT discover qualitatively different dynamical systems.
BPTT mostly finds RNNs with negative LLEs: contracting, predictable dynamics whose gradients vanish over long horizons.
In contrast, the SMT RNN evaluated along Encoder states often has zero or positive surrogate LLEs, i.e., marginal or expansive local dynamics, a regime that is difficult for BPTT to reach due to exploding gradients.
Interestingly, SMT$\rightarrow$DMT RNNs almost always have LLEs near zero, placing them near the edge of chaos, a regime in which recurrent networks have been shown to maximize memory capacity and computational performance~\citep{bertschinger2004edge}.

\newpage

\begin{figure}[t]
    \centering
    \includegraphics[width=1.0\linewidth]{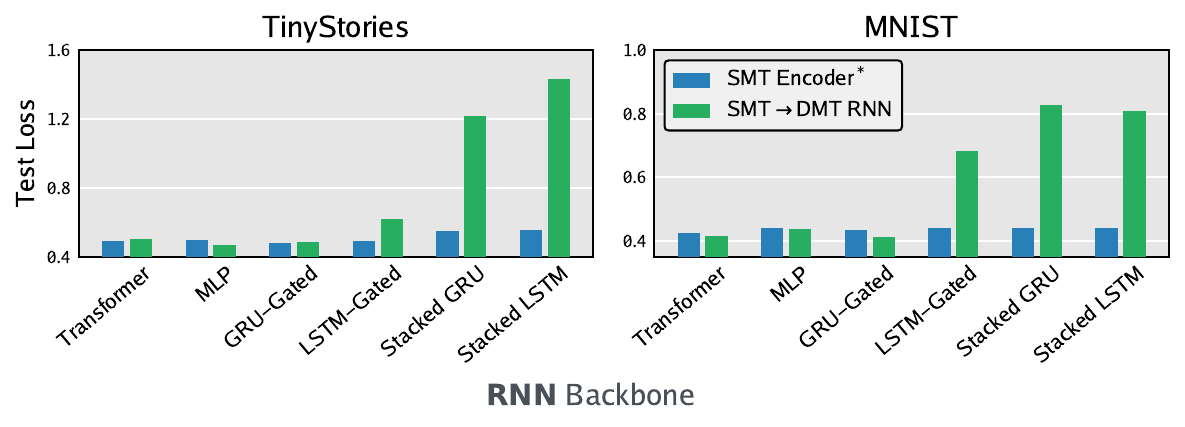}
    \caption{
    \textbf{SMT$\rightarrow$DMT Performance over Architectures.}
    SMT$\rightarrow$DMT is able to train RNNs with a variety of backbones, including GRU/LSTM-Gated models, which are modern RNN architectures with GRU/LSTM style gating applied on the memory updates.
    However, it struggles with vanilla stacked GRUs and LSTMs.
    Figure~\ref{fig:perf_vs_mem_ablation} provides an explanation for why.
    }
    \label{fig:perf_vs_arch}

    \vspace{1em} %

    \includegraphics[width=1.0\linewidth]{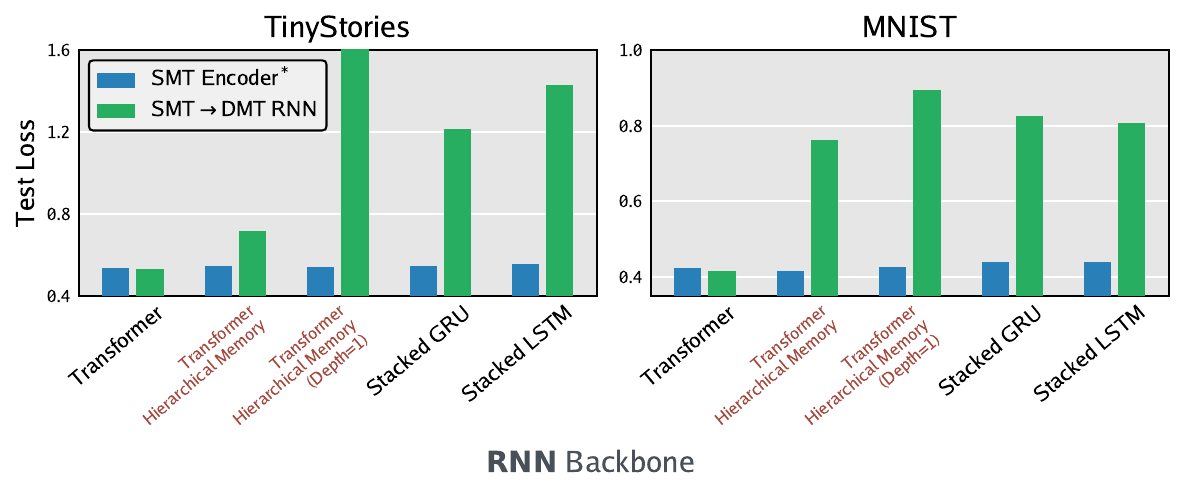}
    \caption{
    \textbf{Why SMT$\rightarrow$DMT fails on Stacked GRU/LSTM.}
    We evaluate SMT and SMT$\rightarrow$DMT on the RNN Transformer architecture, but constrain which memory tokens can view which other memory tokens, by making the attention matrix lower triangular (``Transformer Hierarchical Memory'').
    This mimics a property of stacked GRU/LSTMs, where there is a hierarchical structure in the full memory state.
    We also consider a shallow version of that RNN (``Depth=1''), which mimics the simplicity of the stacked GRU/LSTM's transition function.
    Since SMT$\rightarrow$DMT fails on these architectures as well, it suggests that SMT$\rightarrow$DMT is primarily suited for RNNs which have fully connected memory states with sufficiently expressive transition functions.
    }
    \label{fig:perf_vs_mem_ablation}
\end{figure}

\paragraph{Architectures}
Figure~\ref{fig:perf_vs_arch} shows the performance of SMT$\rightarrow$DMT on TinyStories and MNIST across six RNN architectures (described in Appendix~\ref{sec:appendix_experiment_details_archs}).
SMT$\rightarrow$DMT trains all of them successfully except the vanilla stacked GRU and LSTM.
We hypothesize that this failure stems from two properties of stacked GRUs/LSTMs: their memory state is \textit{hierarchical}, since each layer's state depends only on the layers below it, and their transition function is \textit{shallow}.
To test this, we ablate the Transformer RNN to mimic each property in Figure~\ref{fig:perf_vs_mem_ablation}.
We impose a hierarchical memory structure by making the attention mask between memory tokens lower triangular, so that each memory token can only attend to the tokens preceding it, and we mimic the shallow transition function by reducing the RNN to a single Transformer block.
SMT$\rightarrow$DMT also fails on these ablated RNNs, suggesting that SMT is best suited to RNNs whose memory tokens are fully connected and whose transition functions are sufficiently expressive.
Intuitively, the SMT Encoder is free to produce any memory state, while an RNN with a hierarchical memory structure can only reach states that respect that hierarchy, so the Encoder's memories may lie outside what the RNN can represent.
Extending SMT to hierarchical RNNs may be possible by imposing the same causal structure over the memory tokens in the Encoder, but we leave this to future work.

\begin{figure}
    \vspace{-5pt}
    \centering
    \begin{minipage}{0.45\linewidth}
        \centering
        \includegraphics[width=\linewidth]{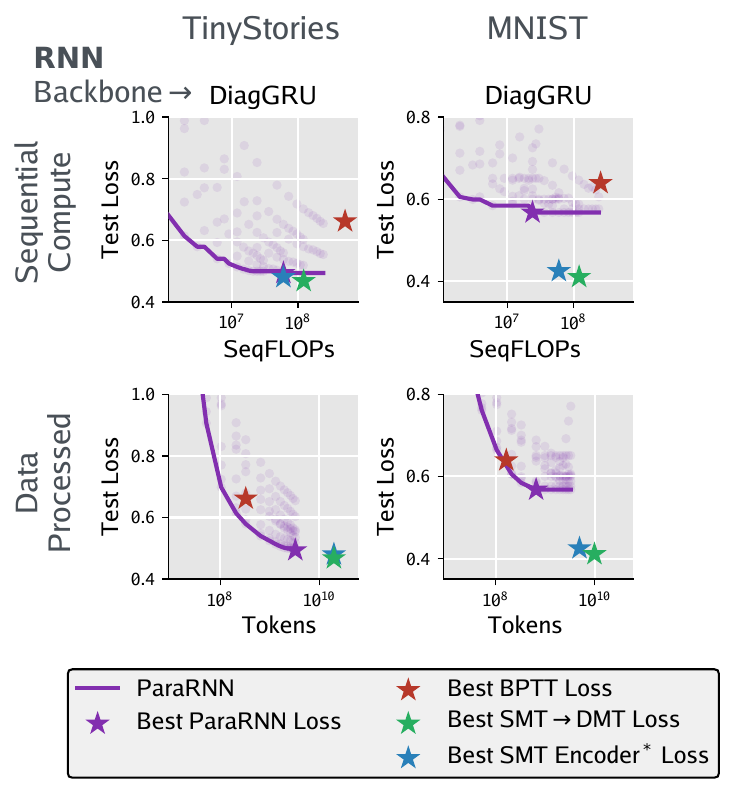}
    \end{minipage}
    \hfill
    \begin{minipage}{0.54\linewidth}
        \caption{
        \textbf{ParaRNN comparison.}
        Here, we evaluate the performance of ParaRNN~\citep{danieli2025pararnnunlockingparalleltraining} (an implementation of DEER~\citep{lim2024parallelizingnonlinearsequentialmodels}) in our setup.
        We find that ParaRNN significantly outperforms BPTT in sequential FLOPs, as expected.
        On TinyStories, ParaRNN is able to match the performance of SMT$\rightarrow$DMT.
        However, ParaRNN lags behind SMT$\rightarrow$DMT on MNIST, likely a result of the long-range dependencies required for the task.
        Because DEER only approximates BPTT, it still struggles with vanishing gradients for long-range dependencies.
        This plot demonstrates that proper long-range credit assignment is as important as parallelization when training sequence models.
        }
        \label{fig:seqflops_pararnn}
    \end{minipage}
\vspace{-5pt}
\end{figure}

\paragraph{Comparison with DEER}
SMT offers two benefits over BPTT: time-parallel training and $\mathcal{O}(1)$ credit path length for long-range memory.
DEER~\citep{lim2024parallelizingnonlinearsequentialmodels} targets only the first.
It parallelizes the RNN forward and backward passes by treating the entire state trajectory as the solution to a system of equations, $\{m_{t+1} - f_\theta(m_t) = 0\}_t$, which it solves with Newton's method.
We compare against ParaRNN~\citep{danieli2025pararnnunlockingparalleltraining}, a scalable implementation of DEER (Figure~\ref{fig:seqflops_pararnn}).
As expected, ParaRNN is far more efficient than BPTT in sequential FLOPs, and on TinyStories it roughly matches SMT$\rightarrow$DMT.
On MNIST, however, which requires remembering strokes from hundreds of timesteps earlier, ParaRNN falls well behind.
This gap arises because DEER only parallelizes (an approximation of) BPTT, and therefore inherits its $\mathcal{O}(T)$ credit path and vanishing gradients.
SMT instead changes the credit assignment itself, which is what long-range memory requires.

\begin{figure}
    \centering
    \includegraphics[width=1.0\linewidth]{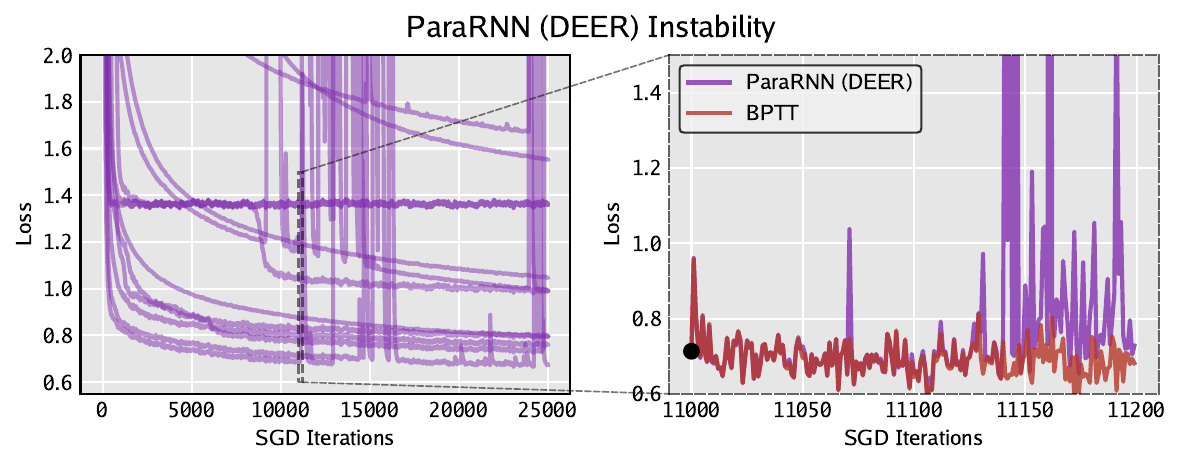}
    \caption{
    \textbf{Instability of ParaRNN.}
    \textbf{Left:}
    Over some runs of training RNNs with DEER~\citep{lim2024parallelizingnonlinearsequentialmodels,danieli2025pararnnunlockingparalleltraining}, we observed severe training instability, as also documented by~\citet{gonzalez2025predictability}.
    \textbf{Right:}
    DEER and BPTT are initialized with the same RNN checkpoint.
    DEER approximates BPTT well up until it becomes unstable, likely because the RNN's dynamical system becomes chaotic~\citep{gonzalez2025predictability}.
    In contrast, SMT enjoys stable optimization which does not depend on the properties of the RNN's dynamics.
    }
    \label{fig:para_instability}
\end{figure}

DEER also has a practical drawback: its inner Newton loop converges only when the RNN's dynamics are predictable (i.e., contracting), and can diverge when they become chaotic~\citep{gonzalez2025predictability}.
We observed this directly (Figure~\ref{fig:para_instability}).
When DEER and BPTT start from the same checkpoint, DEER tracks BPTT closely until training suddenly destabilizes.
This is especially limiting in light of our LLE analysis (Figure~\ref{fig:lle}), as the RNNs that SMT$\rightarrow$DMT discovers lie near the edge of chaos, precisely the regime where DEER's convergence becomes unreliable.
SMT, in contrast, never solves for the RNN's trajectory, so its stability does not depend on the RNN's dynamics.

\newpage

\begin{figure}
    \centering
    \includegraphics[width=1.0\linewidth]{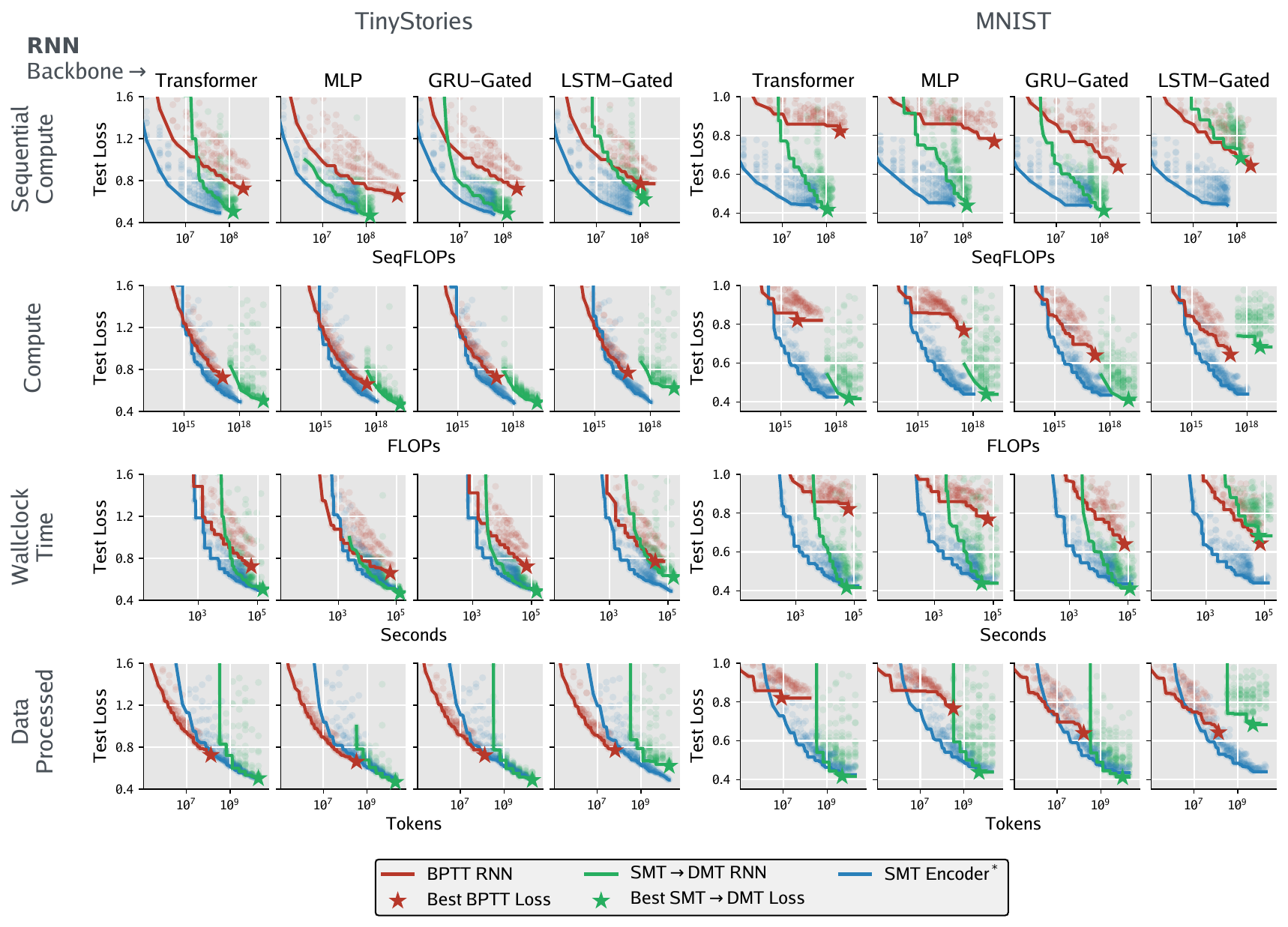}
    \caption{
    \textbf{Sequential Compute, Compute, Wallclock, and Data Efficiency.}
    To complement Figure~\ref{fig:seqflops}, we plot test loss separately against total FLOPs and wall-clock time.
    In FLOPs, we find that SMT$\rightarrow$DMT matches BPTT in TinyStories, but outperforms in MNIST.
    In wallclock time, we find that SMT$\rightarrow$DMT beats BPTT in TinyStories and MNIST.
    }
    \label{fig:seqflops_wallclock}
\end{figure}

\paragraph{Wallclock comparison}
Our main results (Figure~\ref{fig:seqflops}) measure efficiency in sequential FLOPs, which reflects performance on an idealized, infinitely parallel machine.
Since real hardware has finite parallelism, Figure~\ref{fig:seqflops_wallclock} also plots test loss against total FLOPs and wallclock time.
Each run uses a single GPU, so wallclock time is equivalent to GPU hours.
For SMT$\rightarrow$DMT, all metrics account for the resources of the entire pipeline, including training the Encoder, generating its memory targets, SMT, and DMT.
In wallclock time, SMT$\rightarrow$DMT outperforms BPTT on TinyStories and outperforms it by a wide margin on MNIST, mirroring the trends in sequential FLOPs.
In total FLOPs, SMT$\rightarrow$DMT roughly matches BPTT on TinyStories and outperforms it on MNIST.
These two results reflect separate advantages.
The main benefit of parallelization is not FLOP efficiency but the ability to use more FLOPs in the same wallclock time, since BPTT must unroll the RNN one timestep at a time and leaves the GPU underutilized.
This is the same reason Transformers replaced RNNs: time-parallel training unlocked far more compute and thereby enabled scaling~\citep{vaswani2017attention,kaplan2020scaling}.
That SMT$\rightarrow$DMT is \textit{also} more FLOP efficient on MNIST is an additional benefit, likely due to its $\mathcal{O}(1)$ credit path length on tasks requiring long-range memory.

\begin{figure}
    \centering
    \includegraphics[width=1.0\linewidth]{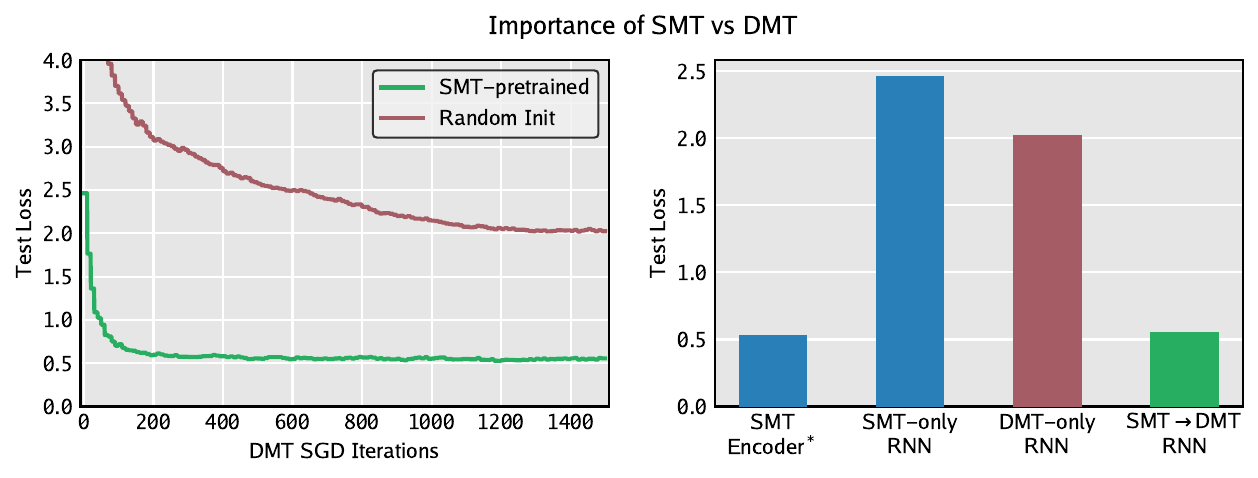}
    \caption{
    \textbf{Both SMT and DMT are critical for performance.}
    On TinyStories, we train a randomly initialized RNN with only DMT, using an SMT-trained Encoder.
    This significantly underperforms initializing with a SMT-pretrained RNN.
    The SMT-pretrained RNN also does not perform well due to drift.
    This suggests that both SMT and DMT are critical for achieving good performance.
    Moreover, only a few iterations of DMT are enough to fix drift, hinting that it is a light finetuning step.
    }
    \label{fig:smt_vs_dmt}
\end{figure}

\paragraph{Importance of SMT vs DMT}
Our full method combines two phases: SMT, which is fully time-parallel, and DMT, which corrects drift but requires unrolling the RNN and thus incurs $\mathcal{O}(T)$ sequential operations (Table~\ref{tab:resources_bigo}).
A natural question is whether both phases are necessary.
Figure~\ref{fig:smt_vs_dmt} isolates each phase on TinyStories.
Using SMT alone yields an RNN that predicts well one step at a time but suffers from drift when rolled out, as small errors compound over the sequence (test loss 2.5).
Conversely, using DMT alone, i.e., training a randomly initialized RNN to imitate the same SMT-trained Encoder, performs significantly worse than starting DMT from an SMT-pretrained RNN (test loss 2.0 vs.\ 0.5).
This suggests that the two phases are complementary: SMT does the heavy lifting of learning what to store in memory, while DMT corrects the errors that arise from rolling the RNN out on its own predictions.
Notably, only a few iterations of DMT are needed to fix drift, so DMT acts as a lightweight finetuning step.
In our runs, 300 DMT iterations following 150K SMT iterations suffice for full performance, with DMT using only 0.67\% of SMT's FLOPs, 0.09\% of its data, and 20\% of its wallclock time (with an unoptimized rollout implementation), and moving the weights only 4\% as far as SMT does.
This is analogous to LLMs, where heavy parallel pretraining is followed by light sequential RL, with both being critical for final performance.
This is important for efficiency, since most of training can then be spent in the parallel SMT phase, while keeping the sequential phase minimal.

\begin{figure}
    \centering
    \begin{minipage}{0.50\linewidth}
        \centering
        \includegraphics[width=\linewidth]{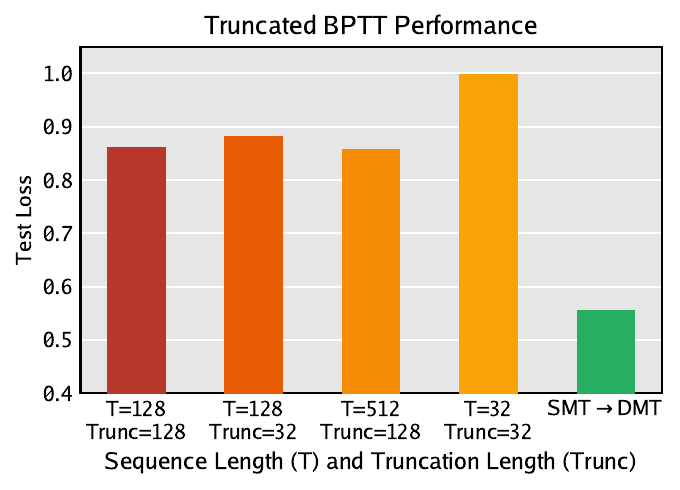}
    \end{minipage}
    \hfill
    \begin{minipage}{0.45\linewidth}
        \caption{
        \textbf{Truncated BPTT Baselines.}
        We evaluate BPTT with different sequence lengths and truncation lengths on TinyStories.
        We find that truncated BPTT does not significantly outperform BPTT in our setup, and is not competitive with SMT$\rightarrow$DMT.
        This is likely because truncation reduces the credit assignment horizon, further hurting BPTT's already poor credit assignment.
        }
        \label{fig:tbptt_baselines}
    \end{minipage}
\end{figure}

\paragraph{Truncated BPTT}
Truncated BPTT (TBPTT)~\citep{williams1990efficient} is a common way to make RNN training cheaper: the sequence is split into chunks, the RNN state is carried forward across chunks, but gradients are only backpropagated within each chunk.
This reduces the sequential cost and memory of each optimization step, at the expense of cutting off credit assignment beyond the truncation length.
Figure~\ref{fig:tbptt_baselines} evaluates BPTT on TinyStories across a range of sequence and truncation lengths.
No configuration of TBPTT significantly outperforms full BPTT, and none is competitive with SMT$\rightarrow$DMT.
This is expected, since TBPTT makes BPTT's credit assignment problem worse rather than better: it explicitly caps the credit path at the truncation length, whereas the bottleneck for BPTT is already its inability to propagate credit over long horizons.
SMT, in contrast, is parallel \textit{and} has an $\mathcal{O}(1)$ credit path length, so it does not face this tradeoff.

\begin{figure}
    \centering
    \begin{minipage}{0.50\linewidth}
        \centering
        \includegraphics[width=\linewidth]{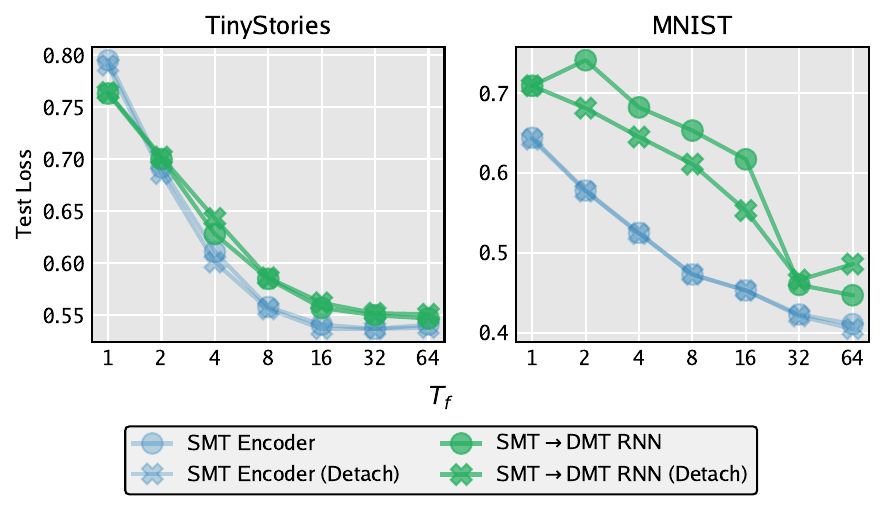}
    \end{minipage}
    \hfill
    \begin{minipage}{0.45\linewidth}
        \caption{
        \textbf{Ablating $T_f$ on TinyStories and MNIST.}
        We find that larger $T_f$ is empirically highly beneficial for SMT.
        Keeping the RNN attached during SMT seems to help only slightly when $T_f$ is large enough.
        However, this provides explicit training signal to the Encoder to make the memory space Markovian, so we keep it by default.
        }
        \label{fig:ablate_Tf}
    \end{minipage}
\end{figure}

\paragraph{Tf Ablation}
The future length $T_f$ controls how far ahead the Decoder must predict from each memory $m_t$, and thus how much predictive information the Encoder is forced to store in memory.
In Section~\ref{sec:detach_rnn_experiments}, we found on the synthetic needle retrieval task that small $T_f$ can be compensated for by joint training, which propagates credit through the RNN dynamics across optimization steps.
Figure~\ref{fig:ablate_Tf} repeats this ablation on TinyStories and MNIST, comparing joint and detached SMT across values of $T_f$.
On these real-world tasks, larger $T_f$ is highly beneficial, suggesting that directly training memories to be predictive of the distant future matters more in practice than credit amortized through the dynamics.
Once $T_f$ is large enough, joint training provides only a slight benefit over detaching the RNN.
We nonetheless use joint training by default, since it explicitly trains the Encoder to produce Markovian memories that the RNN can follow, and set $T_f=64$ in our main experiments.

\begin{figure}
    \centering
    \includegraphics[width=1.0\linewidth]{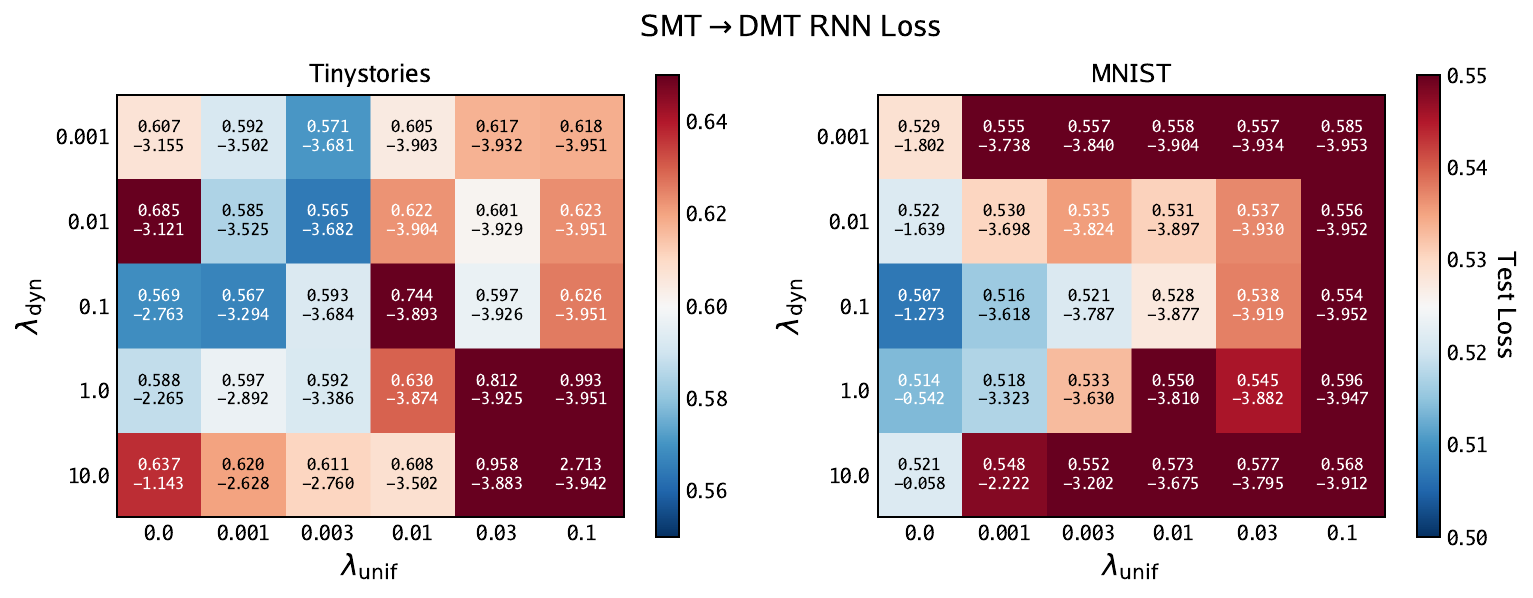}
    \caption{
    \textbf{Sweep of $\lambda_\text{dyn}$ and $\lambda_\text{unif}$.}
    Cell color indicates the RNN test loss for each setting.
    Top number in each cell is the RNN test loss.
    Bottom number in each cell shows the $\mathcal L^\text{unif}$.
    $\mathcal L^\text{unif}$ varies from $0$ (collapsed latent space) to $-4$ (fully uniform latent space).
    }
    \label{fig:lambda_ablation}
\end{figure}

\paragraph{$\lambda_\text{dyn}$ and $\lambda_\text{unif}$ Ablations}
Figure~\ref{fig:lambda_ablation} shows the results of ablating $\lambda_\text{dyn}$ and $\lambda_\text{unif}$.
Results show the optimal RNN performance requires a moderate dynamics loss, paired with a very low uniformity loss.
However, a little uniformity is critical for avoiding memory space collapse.

\begin{figure}[p]
    \centering
    \includegraphics[width=1.0\linewidth]{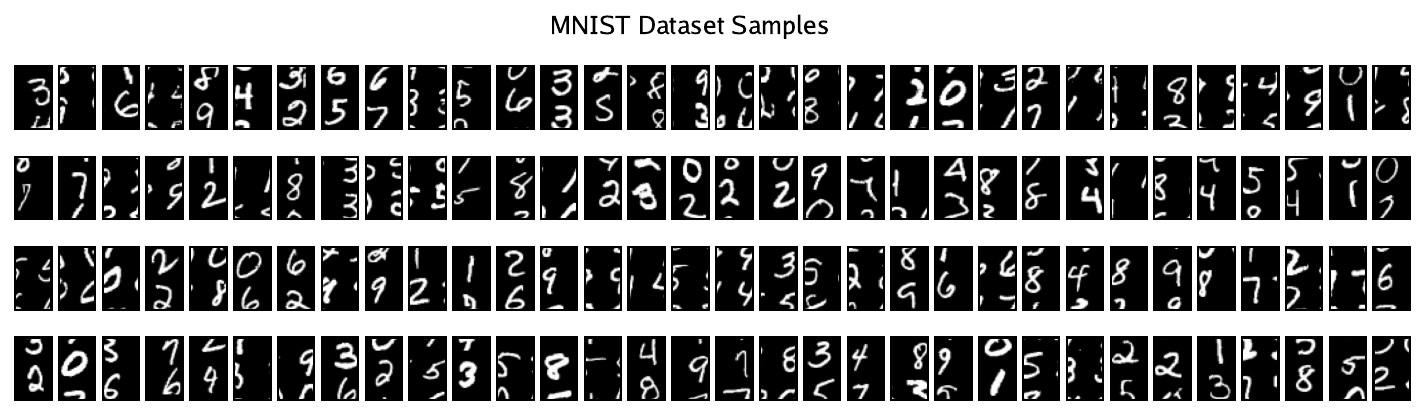}\\[4pt]
    \includegraphics[width=1.0\linewidth]{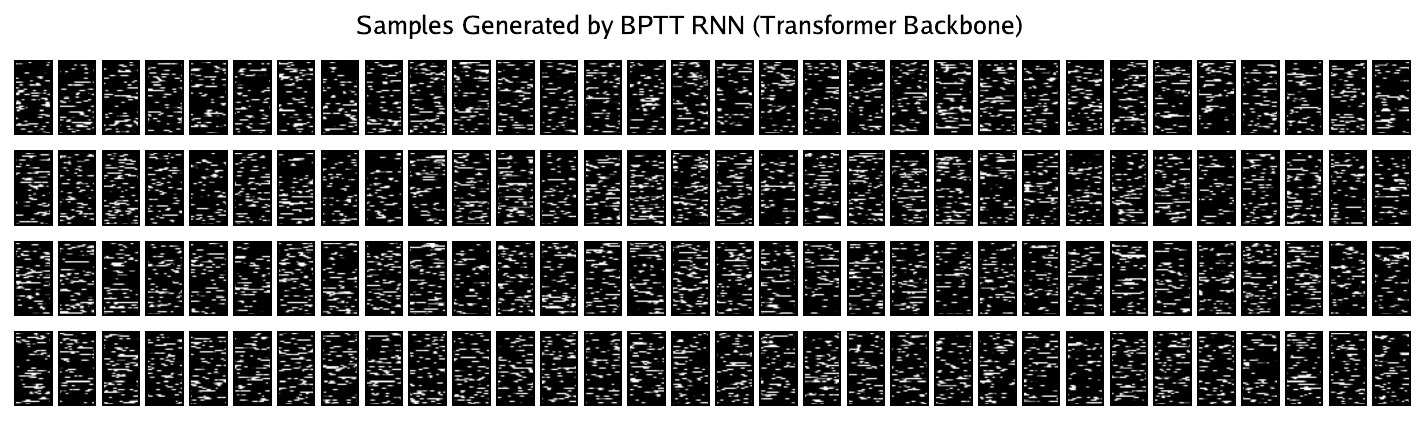}\\[4pt]
    \includegraphics[width=1.0\linewidth]{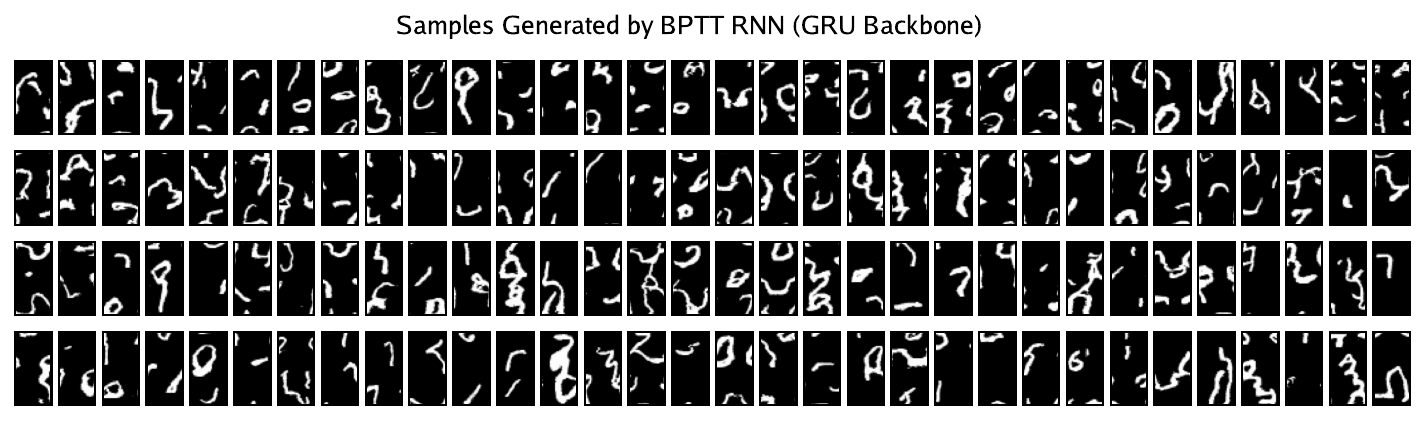}\\[4pt]
    \includegraphics[width=1.0\linewidth]{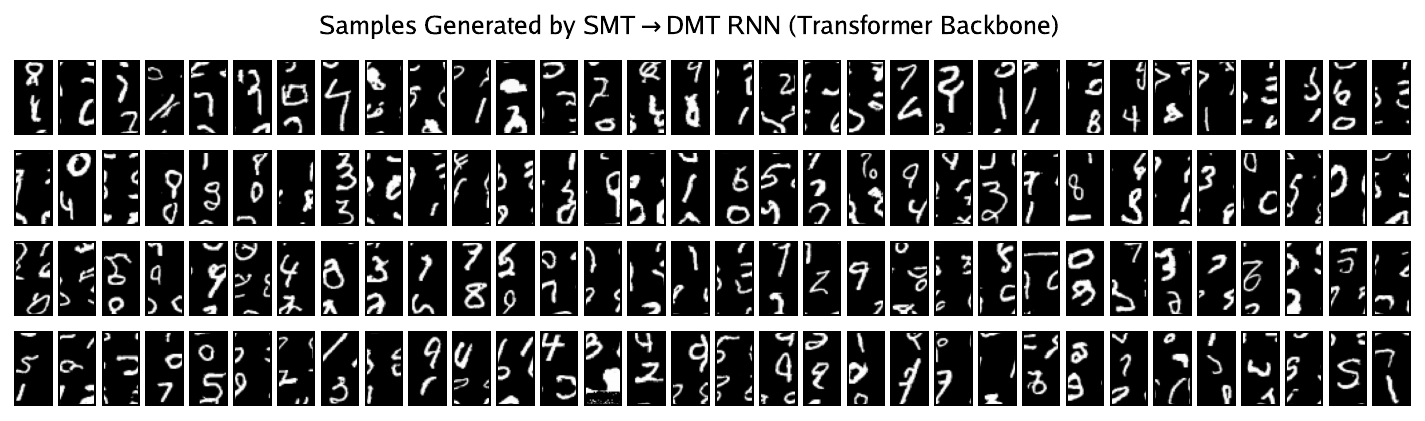}
    \caption{
    \textbf{Additional MNIST Samples.}
    Here we give more examples of samples of MNIST images generated by the various methods.
    SMT$\rightarrow$DMT RNN outperforms BPTT, even when BPTT is applied on a GRU architecture, in processing long-horizon information, which is required for pixel modeling.
    }
    \label{fig:mnist_generations_extra}
\end{figure}

\paragraph{More Pixel Generations}
Figure~\ref{fig:mnist_generations_extra} shows more samples of generations from Figure~\ref{fig:mnist_generations}.
Samples generated by the BPTT RNN (Transformer backbone) seem to only pick up on short range context and act accordingly: either output large streaks of white or black based on the current row.
BPTT RNN (GRU backbone) improves this significantly, but still fails to capture the nuanced structure of digits.
SMT$\rightarrow$DMT RNN (Transformer backbone) is able to capture this structure quite well.

\begin{figure}[p]
    \centering
    \includegraphics[width=1.0\linewidth]{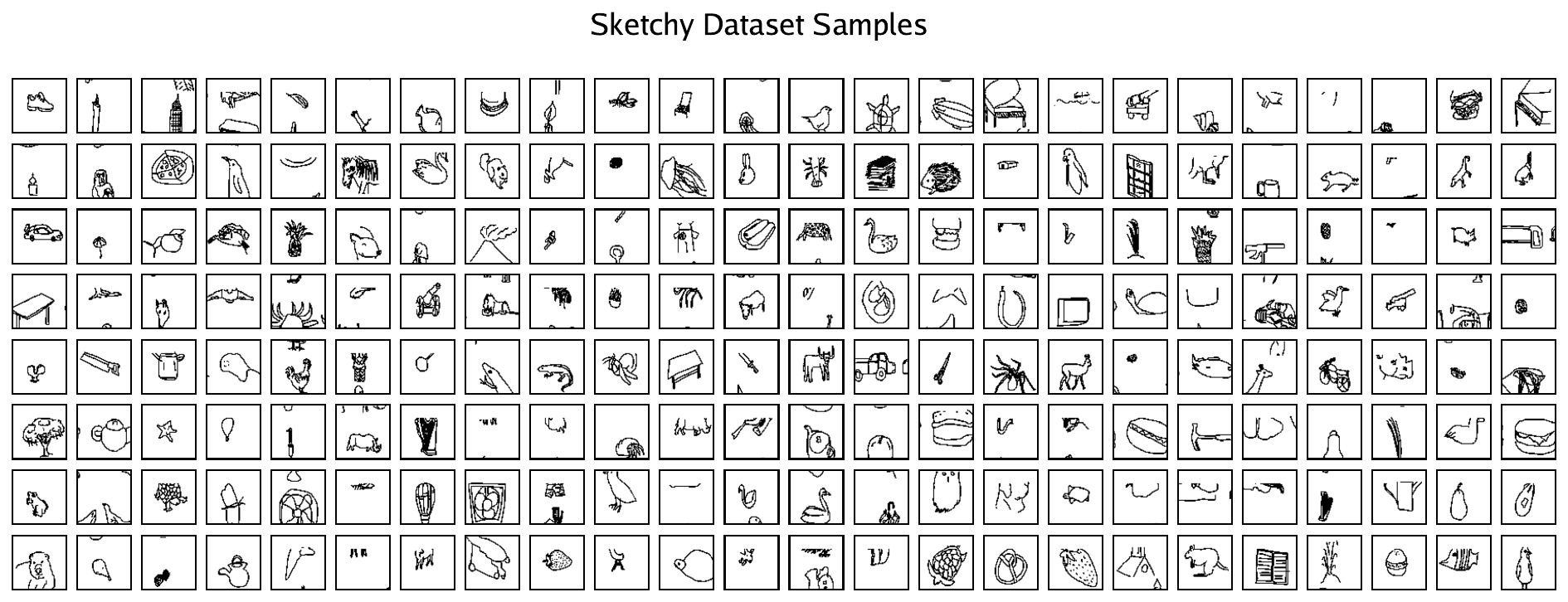}\\[4pt]
    \includegraphics[width=1.0\linewidth]{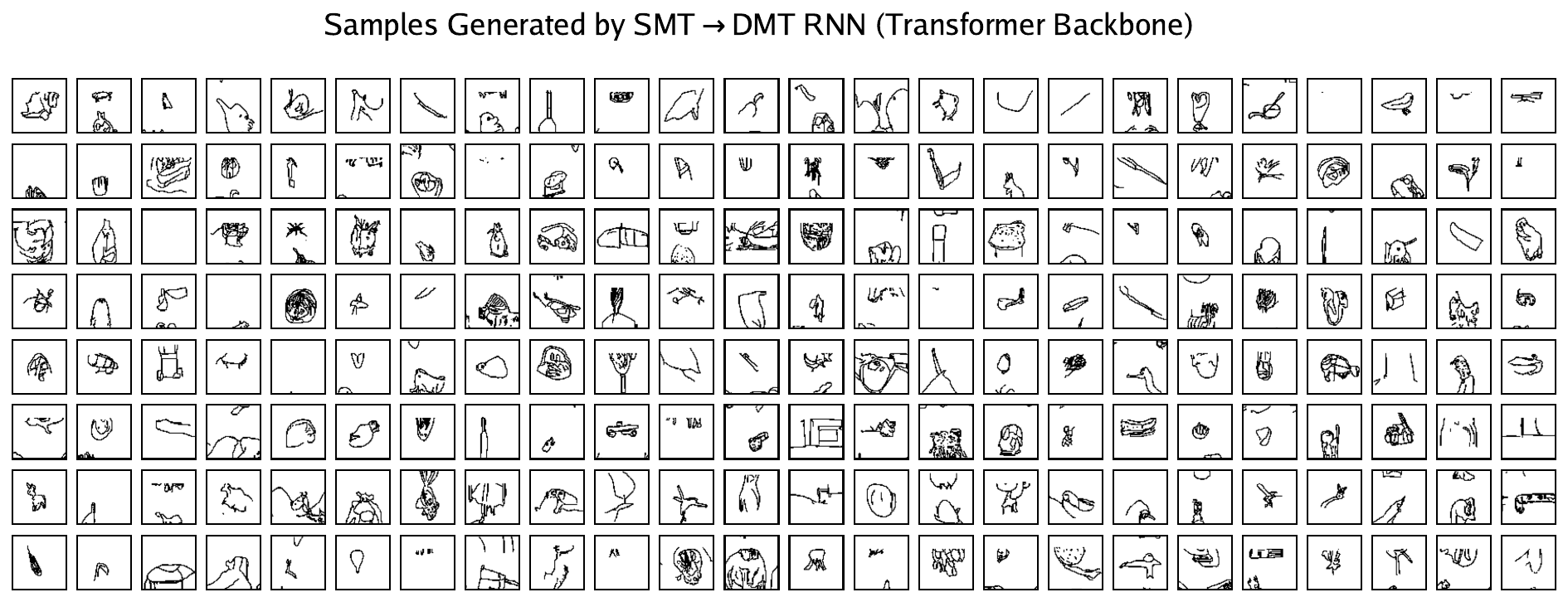}
    \caption{
    \textbf{Additional Sketchy Samples.}
    Here we give more examples of samples of Sketchy images from the dataset and generated by SMT$\rightarrow$DMT.
    Even in this hard sparse domain, SMT$\rightarrow$DMT can capture the overall stroke structure, which requires integrating information over hundreds of pixels.
    }
    \label{fig:sketchy_generations_extra}
\end{figure}

Figure~\ref{fig:sketchy_generations_extra} shows more samples of generations from Figure~\ref{fig:sketchy_generations}.
These generations are often not fully interpretable, but do capture the stroke structure of human-drawn sketches.
Capturing this stroke structure is itself a difficult problem, given the long-horizon nature of pixel sequence modeling.

\begin{figure}
    \centering
    \includegraphics[width=1.0\linewidth]{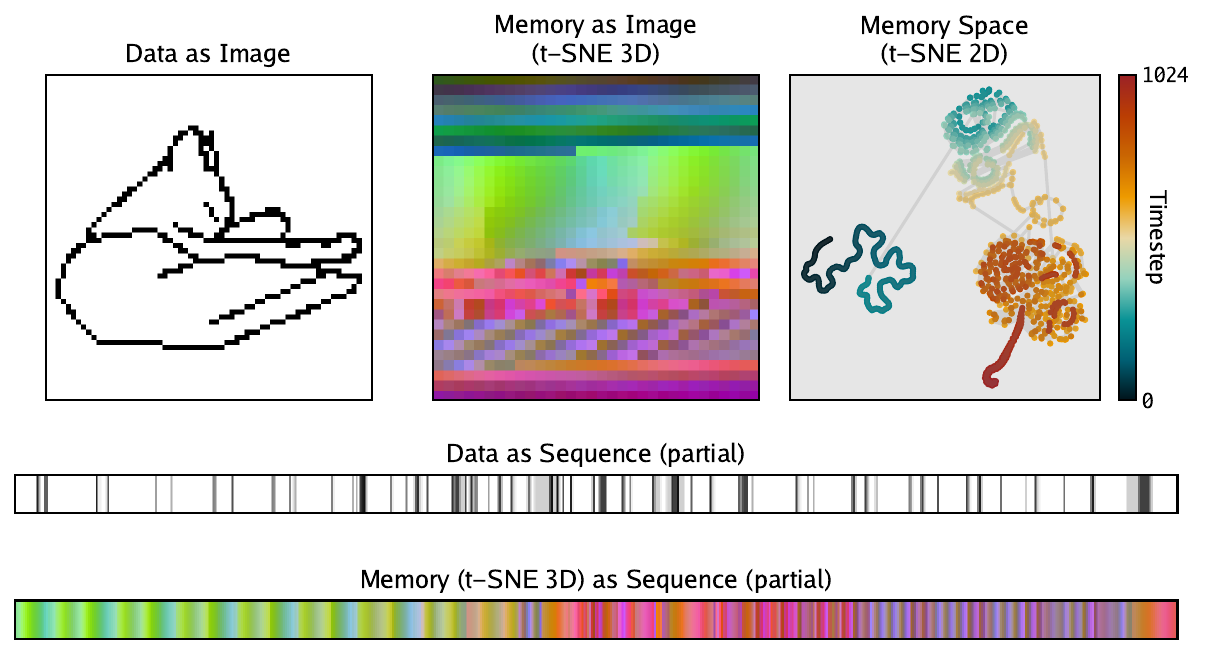}
    \caption{
    \textbf{Analysis on Attneave's Cat.}
    We apply the SMT$\rightarrow$DMT-trained RNN on Sketchy and evaluate it on the classic image of Attneave's cat.
    The RNN reads the image pixel-by-pixel in raster scan order.
    \textbf{Top Left:} Input image presented in its original 2D form.
    \textbf{Top Middle:} 3D t-SNE projection of the RNN memory state, visualized as RGB values over time, showing the evolution of memory throughout sequence processing.
    \textbf{Top Right:} 2D t-SNE projection of the memory state trajectory over time.
    \textbf{Middle:} The same image presented as a flat token sequence.
    From the RNN's perspective, the task resembles modeling a barcode-like sequence, requiring long-range associations between distant tokens and highlighting the difficulty of pixel sequence modeling.
    \textbf{Bottom:} 3D t-SNE projection of the memory state visualized along the flattened sequence.
    }
    \label{fig:cat_read}
\end{figure}

\begin{figure}
    \centering
    \includegraphics[width=1.0\linewidth]{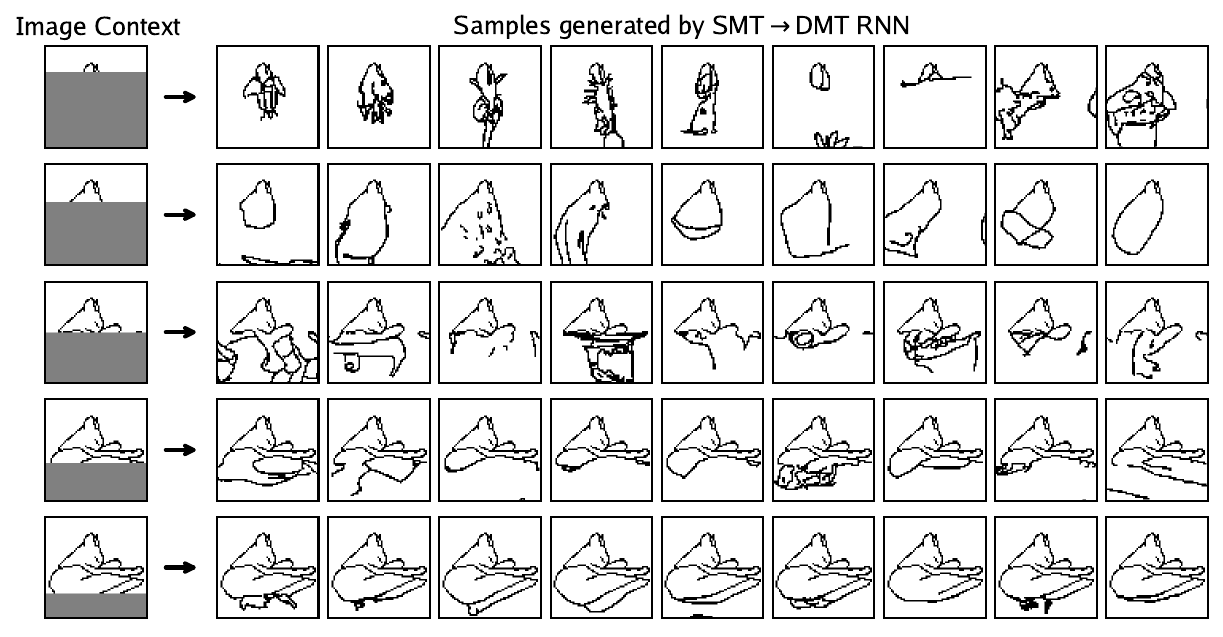}
    \caption{
    \textbf{Generations of Attneave's Cat.}
    We apply the SMT$\rightarrow$DMT-trained RNN on Sketchy and apply it to generate part of the image of Attneave's cat.
    Given more of the image context, the RNN seems to understand the image better and make somewhat more plausible predictions.
    }
    \label{fig:cat_write}
\end{figure}

\begin{figure}[p]
    \centering
    \includegraphics[width=0.7\linewidth]{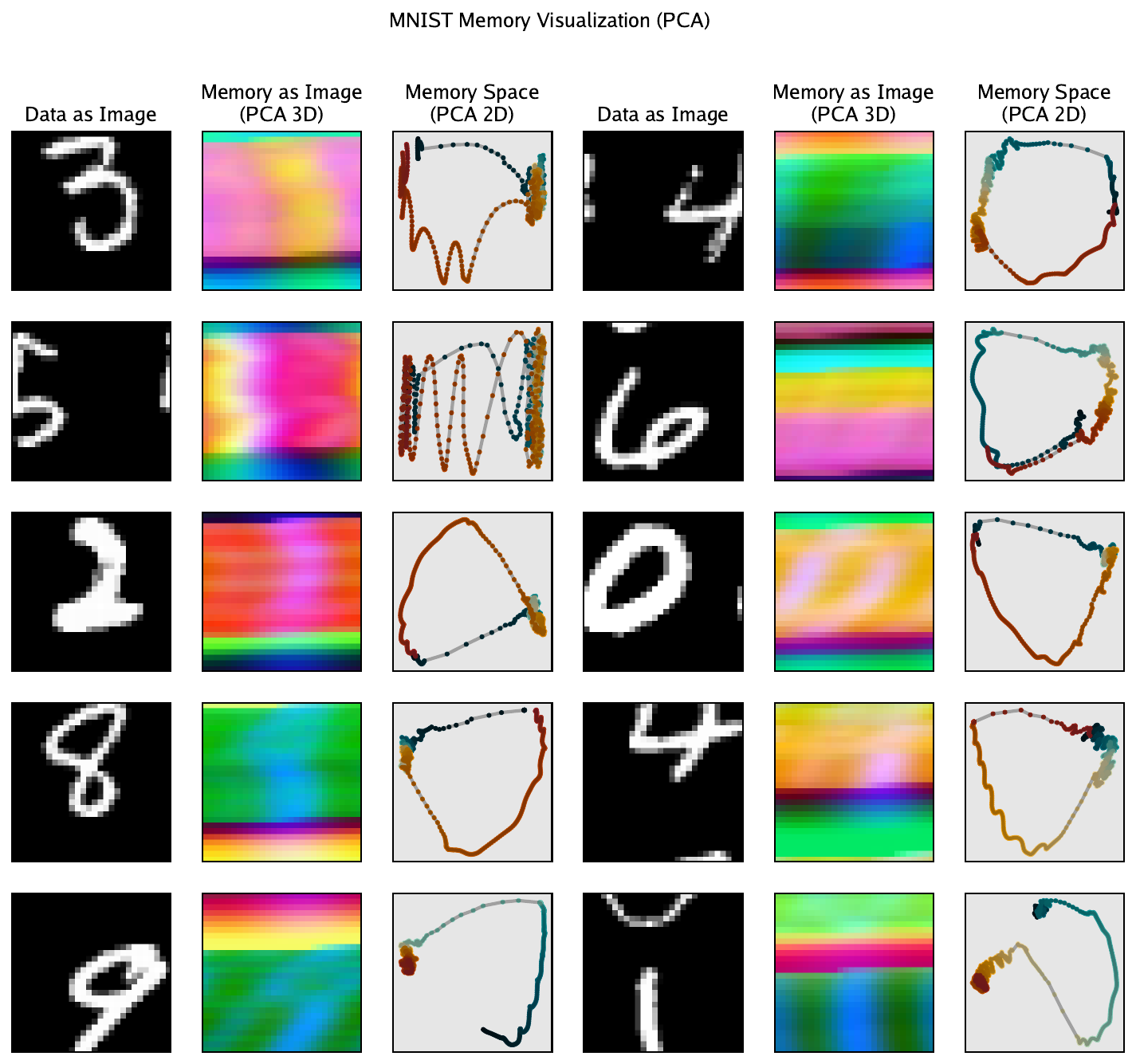}
    \caption{
    \textbf{RNN Memory Evolution on MNIST (PCA)}.
    We analyze the memory evolution of our SMT$\rightarrow$DMT MNIST RNN.
    \textbf{Left}: Input image presented in its original 2D form.
    \textbf{Middle}: 3D PCA projection of the RNN memory state, visualized as RGB values over time, showing the evolution of memory during processing.
    \textbf{Right}: 2D PCA projection of the memory state trajectory over time.
    }
    \label{fig:mnist_memory_pca}
\end{figure}

\begin{figure}[p]
    \centering
    \includegraphics[width=0.7\linewidth]{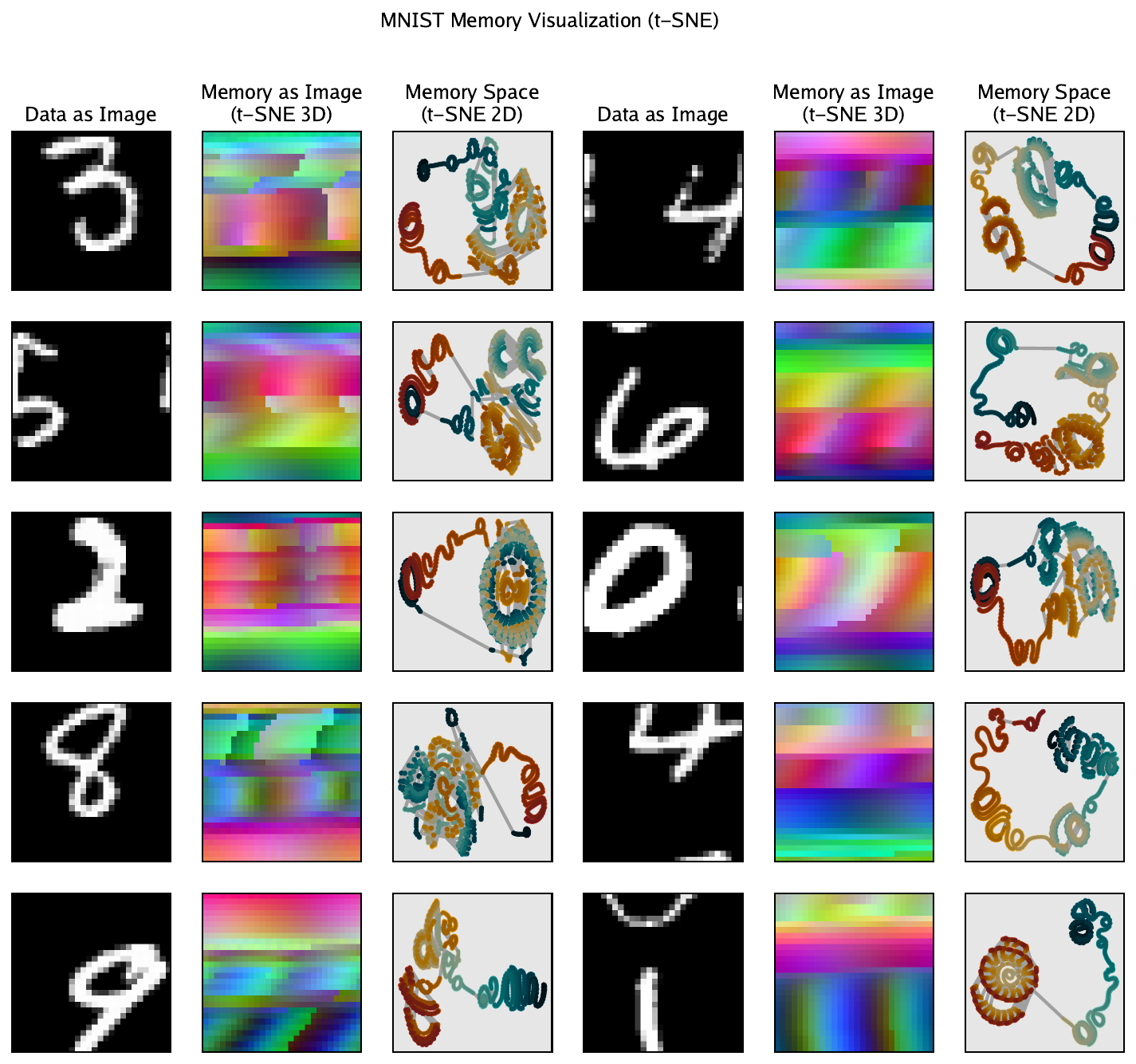}
    \caption{
    \textbf{RNN Memory Evolution on MNIST (t-SNE)}.
    We analyze the memory evolution of our SMT$\rightarrow$DMT MNIST RNN.
    \textbf{Left}: Input image presented in its original 2D form.
    \textbf{Middle}: 3D t-SNE projection of the RNN memory state, visualized as RGB values over time, showing the evolution of memory during processing. 
    \textbf{Right}: 2D t-SNE projection of the memory state trajectory over time.
    }
    \label{fig:mnist_memory_tsne}
\end{figure}

\begin{figure}[p]
    \centering
    \includegraphics[width=1.0\linewidth]{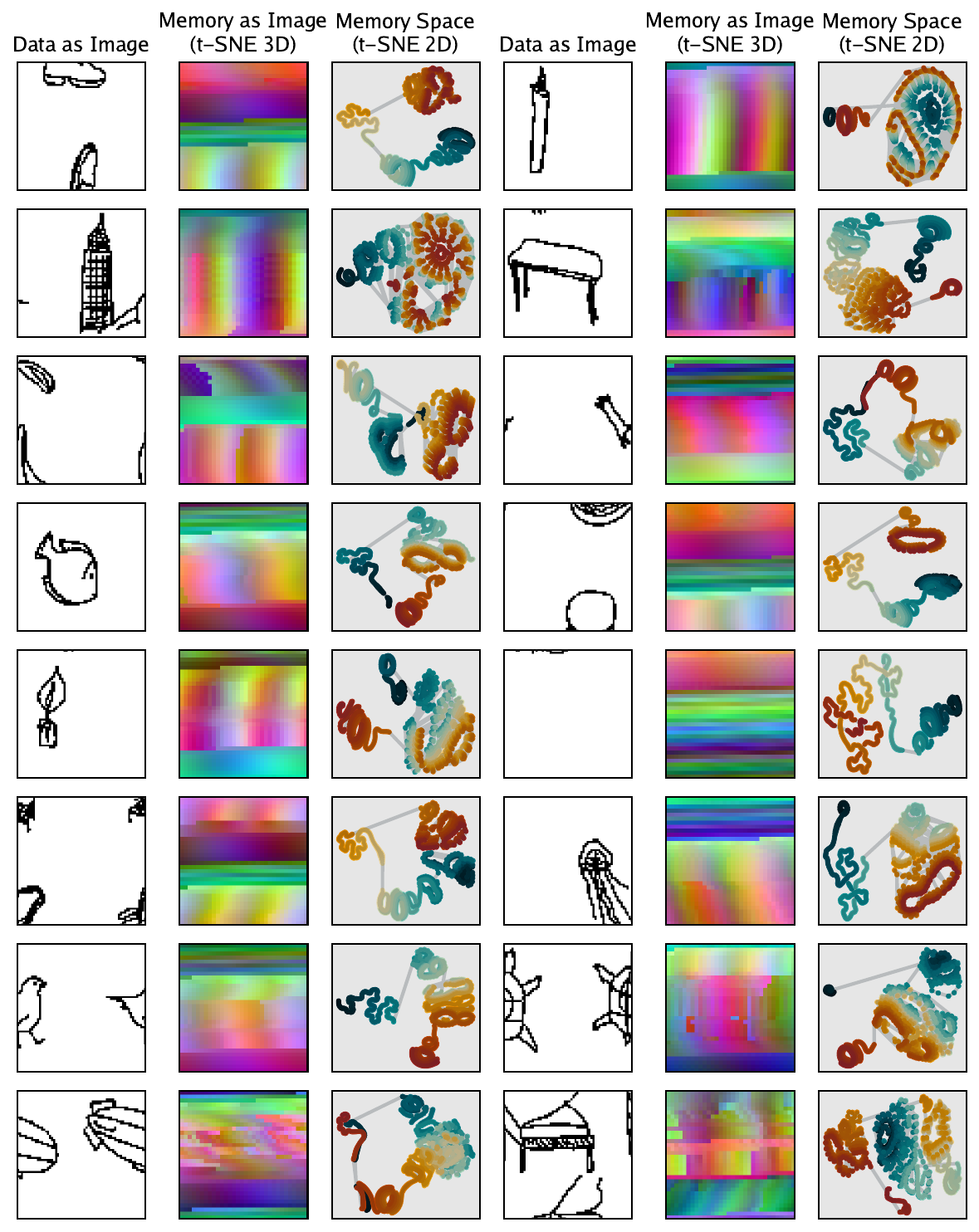}
    \caption{
    \textbf{RNN Memory Evolution on Sketchy (t-SNE)}.
    We analyze the memory evolution of our SMT$\rightarrow$DMT Sketchy RNN.
    \textbf{Left}: Input image presented in its original 2D form.
    \textbf{Middle}: 3D t-SNE projection of the RNN memory state, visualized as RGB values over time, showing the evolution of memory during processing.
    \textbf{Right}: 2D t-SNE projection of the memory state trajectory over time.
    }
    \label{fig:sketchy_memory_tsne}
\end{figure}

\paragraph{Pixel RNN Analysis}
Figure~\ref{fig:cat_read} provides an analysis of our Sketchy RNN as it ``reads'' a sequence corresponding to the classic Attneave's cat image~\citep{attneave1954some}.
The memory sequence does not seem to be fully interpretable, but does show significant structure.
Figure~\ref{fig:cat_write} provides samples of generations when conditioned on partial context of Attneave's cat image.
Figure~\ref{fig:mnist_memory_pca}, \ref{fig:mnist_memory_tsne}, \ref{fig:sketchy_memory_tsne}, show additional analysis of the RNNs on MNIST and Sketchy data.

\begin{figure}[p]
    \centering
    \includegraphics[width=1.0\linewidth]{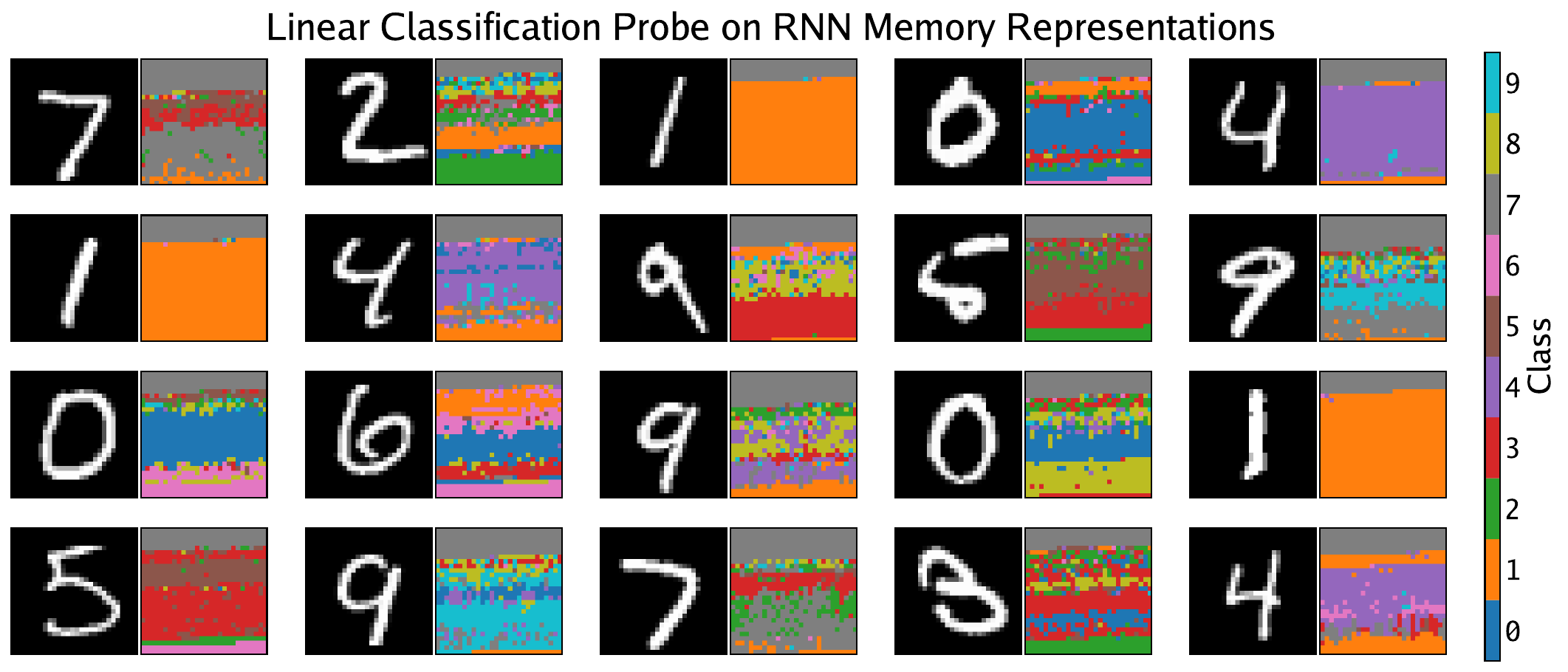}
    \caption{
    \textbf{MNIST Classification Probe on RNN Memory States}.
    We roll out the SMT$\rightarrow$DMT-trained RNN with teacher-forced tokens on MNIST sequences.
    We freeze the RNN and train a linear classification probe on its memory states to investigate how information about the digit class is represented over time.
    Despite seeing the image only one pixel at a time, the RNN develops memory representations that encode semantic information about digit identity.
    The training context length is only 256 tokens, only 33\% of the full 784-pixel image, which explains the degradation in class information beyond a certain horizon.
    }
    \label{fig:mnist_probe_rnn}
\end{figure}

Figure~\ref{fig:mnist_probe_rnn} analyzes how semantic digit information emerges in the RNN as it reads an MNIST digit pixel by pixel.

\section{Compute Resources Used}
\label{sec:appendix_compute_resources}
All individual training runs were conducted on one H200 GPU within 48 hours.
The synthetic experiments comprised more than 375 small-scale training runs, while the real-data experiments required 144 large-scale runs.

\akarshdel{

\vspace{-5pt}
\paragraph{Dynamics and Uniformity Objectives}

We evaluate the effect of sweeping $\lambda_\text{dyn}$, $\lambda_\text{unif}$.
Plot: X-axis: One step $R^2$, y-axis: lossdecoder0 hue: $\lambda_\text{dyn}$, $\lambda_\text{unif}$
no DMT vs DMT vs hot teacher DMT?
\akarsh{DAGGER drift reduction figure}
x axis: drift, y-axis: performance, show trajectory over context before and after DMT
best objective for representation learning? autoencoding or future encoding? next token only? non ppl based? needle in haystack?

\akarsh{three lines:

- Encoder
- RNN with 16 tokens (256 dimensional vector)
- Transformer 

- Sliding window Transformer (16 tokens, )
- Sliding Window Masked Transformer

either equate bits or equate tokens

4096*4 bytes
8*T bytes

}

Plot One step $R^2$ vs Dagger pleateau $R^2$.

visualize memory space and gradient field in 2D of the string copying task and the needle retrieval task. this shows that the memory space is different for different tasks.

- memory markovianness
- memory k-step predictability
- drift
- predictability vs followability
- mechanistic visualizations

- time generalization (computation depth etc)

We find no correlation between one-step $R^2$ and DMT rollout $R^2$, this is interesting future work.

Why is longer ctx z harder to follow?
Spectral norm of jacobian of RNN transition (contraction) does not increase with ctxlen...

We use the following datasets:
Tinystories dataset; character-level language modeling; ascii tokens; vocabsize 256.
Fineweb-10b dataset; language modeling; llama 2 tokens; with vocabsize 32000.
MNIST dataset; pixel-sequence modeling; discrete pixel tokens; vocabsize 256.
CIFAR-10 dataset; pixel-sequence modeling; discrete pixel tokens for each rgb bin; vocabsize 512.
[Protein dataset]; protein sequence modeling; some tokenizer; vocabsize 1024.

Future experiments to consider

Interestingly, we find that SMTDMT RNNs generalize past the teacher in state tracking.
SMT RNNs can be postrained better compared to BPTT
Post training makes RNN more expressive.
Post training allows us to solve state tracking tasks.
Display on parity task with transformer, or on s3 state tracking task.
would be cool to show that RNNs can just generalize past transformer regardless of post training...
}

\akarshdel{
\subsection{Older stuff}
SMT RNNs can outperform SOTA Linear RNNs by using them as teachers

Core BPTT problems: is not time-sequential, training instability, does not do long-term memory
SMT solves all: SMT is time-paralel, stable training, long-range memory

In this section, we compare how SMT compares with BPTT for training various nonlinear RNN architectures.
We compare SMT and BPTT at fixed data/opt steps, and at fixed sequential computations.

We evaluate on the following architectures:
MLP RNN, GRU, Transformer RNN.

Even when BPTT \textit{completely} fails, SMT still works: long string copying, large memory size, bad RNN weight init!

\textbf{Claim 1: } SMT can beat BPTT on most architectures
\textbf{Claim 2: } SMT can create SOTA nonlinear RNNs (by using GRU/LSTM archs)
\textbf{Claim 3: } SMT is more stable

SMT can train beat linear RNNs by using them as a teacher then posttraining.

How does the best SMT RNN compare to best BPTT-LSTM at infinite compute+time?
How does the best SMT RNN compare to linear-RNN at infinite compute+time?
How does the best SMT RNN compare to transformer at infinite compute+time?

\textbf{Baseline pretraining algorithms:}
- BPTT-RNN
- BPTT-LSTM
- SSM
- Transformer

\textbf{Pretraining task:}
- Language Modeling on Fineweb edu

\textbf{Adaptation/Finetuning algorithms:}
- no adaptation
- SFT (BPTT)
- RL
- RandOpt
- z space BPTT

\textbf{Downstream Tasks:}
- Language Modeling
- GSM8K
- MMLU
- LongPPL

\textbf{Evaluations:}
\begin{itemize}
    \item language modeling (perplexity)
    \item perfectly remembering past
    \item needle in the haystack
\end{itemize}

\begin{table}[h]
\centering
\begin{tabular}{l|cccc}
\hline
\textbf{Architecture} & \textbf{BPTT} & \textbf{DEER} & \textbf{SMT} & \textbf{SMT$\rightarrow$DMT}  \\
\hline
MLP RNN &  &  &  &  \\
Vanilla RNN &  &  &  &  \\
GRU RNN &  &  &  &  \\
LSTM RNN &  &  &  &  \\
MemorySetRNN &  &  &  &  \\
\hline
\end{tabular}
\caption{Make this table for different mem sizes for fineweb, tinystories, mnist, cifar, protein folding. Tune LR for each method}
\label{tab:pretrain_architecture}
\end{table}

\begin{table}[h]
\centering
\begin{tabular}{l|ccc}
\hline
\textbf{Pretraining} & \textbf{Posttrain A} & \textbf{Posttrain B} & \textbf{Posttrain C} \\
\hline
BPTT &  &  &  \\
SMT &  &  &  \\
SMT$\rightarrow$ DMT &  &  &  \\
\hline
\end{tabular}
\caption{Task performance for each combination of pretraining and posttraining methods.}
\label{tab:pretrain_posttrain}
\end{table}

Milestones for project:
\begin{enumerate}
    \item milestone 1: beat the same architecture RNN trained with BPTT
    we have this
    \item milestone 2: beat a SOTA LSTM
    i think we can get this, by training SMT on LSTM RNN arch
    \item milestone 2: beat SOTA Linear RNN
    i think we can get this, by making the teacher a linear RNN!
\end{enumerate}
}

\newpage

\newpage

\section{Sequence to Set Reframing}
\label{sec:set_reframing}

As described in Section~\ref{sec:smt}, consider a hypothetical oracle memory-encoding model $\mathcal{Q}$ that takes as input the sequence of tokens and outputs an effective compressed memory.
Here we show that $\mathcal{Q}$ does \textit{not} have to be a recurrent function over the sequence of tokens, but can instead be represented as a permutation-invariant function over a \textit{set} of timestamped tokens.

\vspace{-5pt}
\paragraph{Claim}
Let $\mathbf{x}_\text{seq}=[x_0, x_1, \dots, x_t]$ be the original sequence of tokens.
We define the set $\mathbf{x}_\text{set}=\{(x_0, 0), (x_1, 1), \dots, (x_t, t)\}$.
Assume that $\mathcal{Q}$ is a recurrent function over $\mathbf{x}_\text{seq}$.
In other words,
$$m = \mathcal{Q}(\mathbf{x}_\text{seq})= f(\ldots f(f(m_{\emptyset}, x_0), x_1), \ldots , x_t)$$ with $m_{\emptyset}=\mathbf{0}$ for some function $f$.
For any such $\mathcal Q$, $\exists$ a function $g$ such that
$g(\mathbf{x}_\text{set}) = \mathcal{Q}(\mathbf{x}_\text{seq})$.

\vspace{-5pt}
\paragraph{Proof}
We construct $g$ explicitly. Define $g(\mathbf{x}_\text{set})$ as follows: given the input set $\mathbf{x}_\text{set} = \{(x_0, 0), (x_1, 1), \dots, (x_t, t)\}$, sort the elements in ascending order of their timestamp to recover the sequence $\mathbf{x}_\text{seq} = [x_0, x_1, \dots, x_t]$, then apply $\mathcal{Q}$ to this sequence.

This is well-defined because the timestamps $\{0, 1, \dots, t\}$ are distinct integers, so the sort order is unique. The resulting sequence is identical to the original $\mathbf{x}_\text{seq}$, and therefore $g(\mathbf{x}_\text{set}) = \mathcal{Q}(\mathbf{x}_\text{seq}) = m$.

Moreover, $g$ is permutation-invariant: any permutation of the elements of $\mathbf{x}_\text{set}$ yields the same sorted sequence and thus the same output.

Since $\mathcal{Q}$ was arbitrary, this construction applies to every recurrent $\mathcal{Q}$, completing the proof. $\blacksquare$

\vspace{-5pt}
\paragraph{Implication}
This result implies that any sufficiently expressive permutation-invariant set model can in principle exactly model a recurrent memory function.
Because sets are unordered, time-parallel processing naturally follows.
In particular, Transformer-based architectures can be interpreted as operating over sets of timestamped tokens rather than strictly ordered sequences.

Notably, the proof is constructive: $g$ recovers the sequential computation by sorting the timestamps and implicitly applying the recurrent update rule $f$ up to $t$ times.
Consequently, when implemented with bounded-depth architectures such as Transformers, the required depth may need to scale with sequence length, consistent with prior work on sequential depth and time-parallel training discussed in Section~\ref{sec:related_works}.
Scaling depth with sequence length seems to present a major theoretical limitation.

However, our empirical results suggest that even relatively shallow Transformer encoders can learn highly effective memory representations for both synthetic and natural tasks.
Thus, despite lacking full theoretical expressivity, this sequence-to-set reframing may still provide a practical strategy for memory \textit{pretraining}.
For full expressivity, some light-weight post-training may be required.

\newpage
\section{Encoder Markovian Training}
\label{sec:markovian_encoder_proof}
SMT consists of two primary objectives: future predicting with $\mathcal{L}^{\text{dec}}$ and dynamics modeling with $\mathcal{L}^{\text{dyn}}$.
The dynamics objective serves two purposes: (1) training the RNN to predict the next memory state from the current one, and (2) encouraging the encoder to produce memory states that are predictable from one another, i.e. approximately Markovian.
In this section, we show that the predictive state objective $\mathcal{L}^{\text{dec}}$ alone is sufficient for learning Markovian memories, implying that $\mathcal{L}^{\text{dyn}}$ is theoretically unnecessary, though still practically useful.

\vspace{-5pt}
\paragraph{Claim}
Let $\mathbf{x} = [x_0, x_1, \dots, x_T]$ and $\mathbf{y} = [y_0, y_1, \dots, y_T]$
be input and output sequences. For each timestep $t$, define
\[
  \mathbf{x}_t^{\text{ctx}} = [x_0, \dots, x_t], \qquad
  \mathbf{x}_t^{\text{fut}} = [x_{t+1}, \dots, x_T], \qquad
  \mathbf{y}_t^{\text{fut}} = [y_t, \dots, y_T],
\]
with memory state $m_t = \mathcal{E}_\phi(\mathbf{x}_t^{\text{ctx}})$ and reconstructed
future $\hat{\mathbf{y}}_t^{\text{fut}} = \mathcal{D}_\psi(m_t, \mathbf{x}_t^{\text{fut}})$.
If $m_t$ is an \textit{optimal} minimal sufficient statistic of $\mathbf{x}_t^{\text{ctx}}$ for predicting $\mathbf{y}_t^{\text{fut}}$ given $\mathbf{x}_t^{\text{fut}}$ at \textit{every} $t$, then the memory sequence $(m_t)$ is \textbf{Markovian}:
\[
  m_{t+1} \perp\!\!\!\perp \mathbf{x}_t^{\text{ctx}} \;\Big|\; m_t,\, x_{t+1}.
\]
\vspace{-5pt}
\paragraph{Proof}
By optimality of $m_t$, it is a minimal sufficient statistic of $\mathbf{x}_t^{\text{ctx}}$
for predicting $\mathbf{y}_t^{\text{fut}}$ given $\mathbf{x}_t^{\text{fut}}$:
\[
  H\!\bigl(\mathbf{y}_t^{\text{fut}} \mid m_t, \mathbf{x}_t^{\text{fut}}\bigr)
  = H\!\bigl(\mathbf{y}_t^{\text{fut}} \mid \mathbf{x}_t^{\text{ctx}}, \mathbf{x}_t^{\text{fut}}\bigr).
\]
Note that $\mathbf{y}_{t+1}^{\text{fut}} \subseteq \mathbf{y}_t^{\text{fut}}$ and
$\mathbf{x}_{t+1}^{\text{fut}} \subseteq \mathbf{x}_t^{\text{fut}}$, so optimality of $m_t$
at time $t$ implies it is also sufficient for $\mathbf{y}_{t+1}^{\text{fut}}$ given
$\mathbf{x}_{t+1}^{\text{fut}}$:
\[
  H\!\bigl(\mathbf{y}_{t+1}^{\text{fut}} \mid m_t, x_{t+1}, \mathbf{x}_{t+1}^{\text{fut}}\bigr)
  = H\!\bigl(\mathbf{y}_{t+1}^{\text{fut}} \mid \mathbf{x}_t^{\text{ctx}}, x_{t+1}, \mathbf{x}_{t+1}^{\text{fut}}\bigr)
  = H\!\bigl(\mathbf{y}_{t+1}^{\text{fut}} \mid \mathbf{x}_{t+1}^{\text{ctx}}, \mathbf{x}_{t+1}^{\text{fut}}\bigr).
\]
By optimality of $m_{t+1}$, it is a minimal sufficient statistic of
$\mathbf{x}_{t+1}^{\text{ctx}}$ for predicting $\mathbf{y}_{t+1}^{\text{fut}}$ given
$\mathbf{x}_{t+1}^{\text{fut}}$. Minimality means $m_{t+1}$ retains no information
from $\mathbf{x}_{t+1}^{\text{ctx}}$ beyond what is predictively necessary. Since
$(m_t, x_{t+1})$ already constitutes a sufficient statistic for this same
prediction task---as shown above---minimality of $m_{t+1}$ forces it to be a
function of $(m_t, x_{t+1})$:
\[
  m_{t+1} = f(m_t,\, x_{t+1})
\]
for some measurable $f$. Therefore $m_{t+1}$ is determined entirely by
$(m_t, x_{t+1})$, and conditioning on these renders it independent of all
earlier context:
\[
  H(m_{t+1} \mid m_t,\, x_{t+1}) = 0,
\]
which is equivalent to $m_{t+1} \perp\!\!\!\perp \mathbf{x}_t^{\text{ctx}} \mid m_t, x_{t+1}$.
Hence $(m_t)$ is Markovian. $\blacksquare$

\vspace{-5pt}
\paragraph{Implication}
This result establishes that under ideal conditions---sufficient encoder and decoder
capacity, infinite future horizon, and exact optimization---the memory states learned
by $\mathcal{E}_\phi$ form a Markov chain driven only by the previous state and the
incoming token. In other words, the encoder \textit{implicitly} learns a Markovian memory representation:
$m_{t+1}$ can be predicted from only $(m_t, x_{t+1})$.

In practice, finite capacity and approximate optimization relax this property, leaving
$m_{t+1}$ with residual dependence on $\mathbf{x}_t^{\text{ctx}}$ beyond $(m_t, x_{t+1})$.
This gap motivates jointly training the dynamics loss $\mathcal{L}^{\text{dyn}}$
alongside $\mathcal{L}^{\text{dec}}$ to explicitly encourage Markovian structure in the learned memory sequence.

Relatedly, \citet{teoh2025nextlatentpredictiontransformerslearn} provide a proof that one-step RNN dynamics with a $T_f=1$ encoder also induce a predictive state memory representation.

\akarshdel{
This stuff will get deleted or moved to appendix. I think this is obvious for RNNs.

\subsection{Interpretations}

\paragraph{SMT is a generalization of Sliding Window Attention}
Sliding Window Attention (SWA) is technically a fixed-memory-size RNN which updates its memory tokens by using a queue/circular buffer: it adds new tokens at the front and removes the least recent tokens from the back.
Normal Attention can be seen a growing-memory-size RNN which updates its memory by using a list: it adds new tokens at the front, without removing anything. 
Our SMT training paradigm can be seen as a generalization of the SWA and a fixed-memory-size compression of normal attention.
Essentially instead of hardcoding the update function to be a queue operation over token space, we learn the update function over some memory space, trained with ground truth memory labels from a encoder.

Transformers aren't doing hard time-compression!! RNNs are!!

Additionally, it can be useful to compare BPTT, SMT, and DMT from an reinforcement learning (RL) perspective.
BPTT can be seen as a form of policy gradients, where the RNN is being rolled out on-policy, and the gradient credit assignment is done by BPTT instead of REINFORCE~\citep{}.
SMT can be seen as behavior cloning of an expert in the memory space.
DMT can be seen as the DAgger algorithm, on-policy imitation learning of an expert in the memory space.

Let us define the "optimal RNN" as the one that at any point in time, t, has represented all past information necessary for optimal decision making at time t. That is, the RNN's hidden state is a sufficient statistic of the history w.r.t. the decision making problem. The optimal RNN, coupled with the optimal decoder, achieves the max possible performance on the decision making task.
Under what conditions does our training method recover the optimal RNN?
Suppose the sequence we are modeling is a Markov process of order n (dependencies only go n-back).
Condition 1: The encoder must have ctxlen + futlen $>=$ n?
What is our savings compared to the QueueRNN of ctxlen=n?
The memory size of the QueueRNN is O(n).
The memory size of our RNN is something like O(n/s), where s is the "time compressibility" of the sequence. For example, parity checks have s=n since you can convert an n length sequence into a length 1 sequence updated n times.
}

\end{document}